\documentclass[10pt,twocolumn,letterpaper]{article}

\usepackage{cvpr}              % To produce the CAMERA-READY version
\usepackage[dvipsnames]{xcolor}

\usepackage{subcaption}
\usepackage{multirow}
\usepackage{algorithm}
\usepackage{algorithmic}
\usepackage{amssymb}
\usepackage{tcolorbox}
\usepackage{booktabs}
\usepackage{pifont}
\usepackage{color}
\usepackage{algorithm}
\usepackage{amsmath}
\usepackage{algorithmic}
\usepackage{makecell}
\newcommand{\cmark}{\ding{51}} % check mark
\newcommand{\xmark}{\ding{55}} % cross mark
\usepackage{newfloat}
\usepackage{listings}
\usepackage{afterpage}
\usepackage{colortbl} 
\definecolor{cvprblue}{rgb}{0.21,0.49,0.74}
\usepackage[pagebackref,breaklinks,colorlinks,citecolor=cvprblue]{hyperref}

\def\paperID{xxxx} % *** Enter the Paper ID here
\def\confName{CVPR}
\def\confYear{2024}

\title{Can LVLMs Obtain a Driver's License? A Benchmark Towards Reliable AGI \\for Autonomous Driving}
\author{\textbf{
Yuhang Lu$^{1,}$\thanks{These authors contributed equally.}, 
Yichen Yao$^{1,}$\footnote[1]{}, 
Jiadong Tu$^{1,}$\footnote[1]{}, 
Jiangnan Shao$^{1,}$\footnote[1]{}, 
Yuexin Ma$^{1,}$\thanks{Corresponding author.} , 
Xinge Zhu$^{2,}$\footnote[2]{}
}\\ 
$^{1}$ ShanghaiTech University
$^{2}$ The Chinese University of Hong Kong\\
{\tt\small \{luyh2, yaoych2023, tujd2023, shaojn2023, mayuexin\}@shanghaitech.edu.cn} \\ {\tt\small zhuxinge123@gmail.com}}

\begin{document}
\maketitle

%%%%%%%%% ABSTRACT
\begin{abstract}
   Large Vision-Language Models (LVLMs) have recently garnered significant attention, with many efforts aimed at harnessing their general knowledge to enhance the interpretability and robustness of autonomous driving models. However, LVLMs typically rely on large, general-purpose datasets and lack the specialized expertise required for professional and safe driving. Existing vision-language driving datasets focus primarily on scene understanding and decision-making, without providing explicit guidance on traffic rules and driving skills, which are critical aspects directly related to driving safety. To bridge this gap, we propose IDKB, a large-scale dataset containing over one million data items collected from various countries, including driving handbooks, theory test data, and simulated road test data. Much like the process of obtaining a driver's license, IDKB encompasses nearly all the explicit knowledge needed for driving from theory to practice. In particular, we conducted comprehensive tests on 15 LVLMs using IDKB to assess their reliability in the context of autonomous driving and provided extensive analysis. We also fine-tuned popular models, achieving notable performance improvements, which further validate the significance of our dataset. The
project page can be found at: \url{https://4dvlab.github.io/project_page/idkb.html}

%Large Vision-Language Models (LVLM) has become a popular focus in AI, gaining significant attention for its strong performance across various domains. When combined with autonomous driving algorithms, LVLMs can enhance the performance and interpretability of these systems. However, LVLMs typically rely on large, general-purpose datasets and lack the specific expertise needed for the driving domain, requiring targeted fine-tuning. Existing visual-language autonomous driving datasets primarily focus on scenario understanding and decision-making, without offering explicit guidance on safe driving rules and structured learning. To address this gap, we developed the IDKB dataset, inspired by the human process of obtaining a driver's license. The dataset comprises over one million data items, including driving handbooks and test data from various countries, as well as CARLA-generated road scenario data. We also evaluated 15 LVLMs using this dataset, and the significant performance improvements after fine-tuning underscore the value of our approach.

\end{abstract}

%%%%%%%% BODY TEXT
\section{Introduction}
\label{sec:intro}
In recent years, Large Vision-Language Models (LVLMs) \cite{achiam2023gpt, bai2023qwen, liu2024visual} have emerged as powerful tools in AI, showcasing impressive capabilities in areas such as visual dialogue and document understanding. Building on the general knowledge of LVLMs, some approaches \cite{xu2023drivegpt4,mao2023language,sima2023drivelm} have leveraged these models to enhance the efficiency, robustness, and interpretability of autonomous vehicles, addressing the intricate challenges of autonomous driving in open world. However, LVLMs are often trained on vast and generic datasets, lacking the specialized expertise required for the driving domain. This gap in domain-specific knowledge can lead to potential inaccuracies when these models are applied to self-driving systems, where precision and reliability are paramount.

\begin{figure}[t]
  \includegraphics[width=1\linewidth]{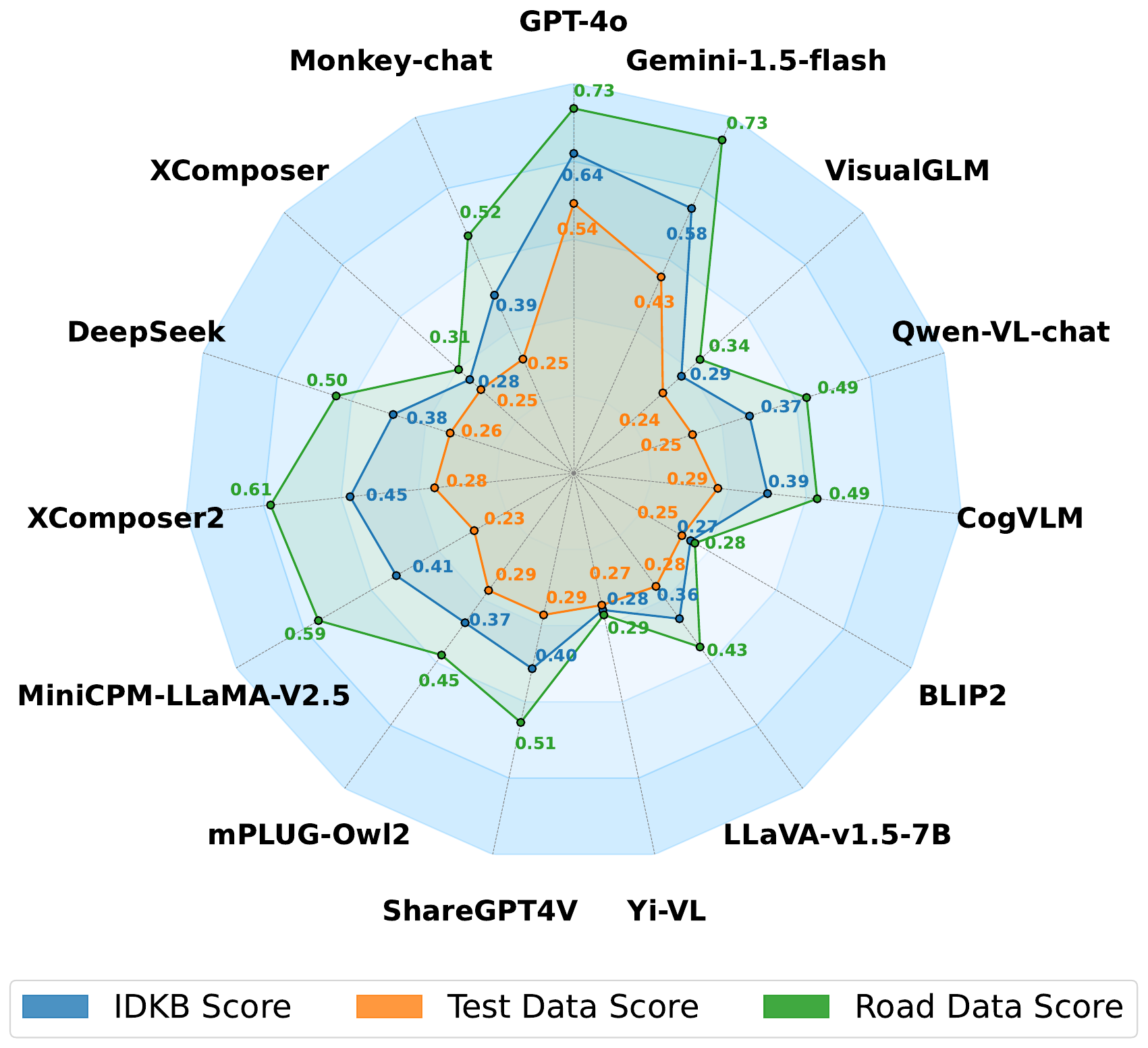}
  \caption{Performance of 15 representative Large Vision-Language Models on IDKB, evaluated by three driving knowledge understanding metrics.}
\label{fig:teaser}
\end{figure}

To address this issue, many vision-language driving datasets \cite{ qian2024nuscenes, sima2023drivelm, park2024vlaad, kim2018textual, malla2023drama, tian2024drivevlm, li2024automated} towards LVLM fine-tuning have been developed. Most of these datasets simply add textual annotations to traffic images from existing datasets, which limits the complexity and diversity of the scenarios they cover. While a few datasets~\cite{tian2024drivevlm, li2024automated} are specifically collected and annotated with more challenging driving scenarios, they still primarily focus on tasks like scene perception and decision-making, rather than providing structured driving knowledge. As a result, models built on these datasets can only implicitly learn driving knowledge through the supervision of driving decisions. However, this approach differs significantly from how humans learn to drive, which involves studying driving instructions, traffic laws, driving rules, driving skills, and methods for handling emergency situations. Consequently, these models often lack a comprehensive understanding of driving knowledge, leading to unstable and unreliable performance in real-world applications.

To this end, we propose \textbf{I}ntelligent \textbf{D}riving \textbf{K}nowledge \textbf{B}ase (IDKB), the first large-scale vision-language dataset dedicated to professional driving knowledge and experience. Typically, humans learn to drive systematically by studying driving materials, taking theory tests, and practicing on the road. To enable LVLMs to effectively ``earn a driver's license'' and guarantee their driving safety, we compiled an extensive collection of driving handbooks and test questions from various countries, covering traffic laws, rules, driving techniques, and crisis management skills. In addition to theoretical data, we generated practical data by simulating diverse road scenarios in CARLA~\cite{dosovitskiy2017carla}, including variations in weather, lighting, traffic conditions, and more. This comprehensive effort resulted in a dataset of over 1 million data entries with various formats, spanning 15 countries, 9 languages, and 4 vehicle types. By capturing diverse driving regulations and practices from various regions, our dataset provides a solid foundation for thoroughly evaluating LVLMs and enhancing their ability to acquire safe and efficient driving capabilities.

Based on IDKB, we conducted an extensive evaluation for 15 existing LVLMs to assess their degree of mastery in driving knowledge and skills. As shown in Fig.~\ref{fig:teaser}, the evaluated LVLMs generally lacked strong driving domain knowledge, underscoring the need for fine-tuning with high-quality, structured, and diverse driving knowledge data for effective application in autonomous driving.

We also fine-tuned several of these models using our dataset, and the experimental results demonstrate that explicit and structured driving knowledge significantly enhances the performance of LVLMs, leading to more effective and accurate outcomes. Our findings highlight the importance of incorporating specialized, domain-specific knowledge into LVLMs to better equip them for the complex and safety-critical task of autonomous driving.

Our key contributions can be summarized as follows:
\begin{itemize}
\item We introduce IDKB, the first large-scale vision-language dataset explicitly containing both driving theory and practical knowledge.
\item We evaluate 15 existing LVLMs on our dataset and provide a comprehensive analysis of their driving abilities.
\item We offer fine-tuned, open-source LVLMs trained on our dataset, which possess enhanced professional driving expertise.
\end{itemize}

\section{Related Work}
\label{sec:related}
\subsection{LVLMs for Autonomous Driving}

Recently, research on Large Vision-Language Models (LVLMs) has surged, with multimodal models such as GPT-4V~\cite{achiam2023gpt}, Qwen~\cite{bai2023qwen}, and LLaVA~\cite{liu2024visual} demonstrating strong performance across a wide range of general tasks. Leveraging these generalized capabilities, several approaches have begun to integrate LVLMs with autonomous driving algorithms to enhance self-driving car performance and interpretability. For example, DriveGPT4~\cite{xu2023drivegpt4} processes multimodal input data and generates both text responses and vehicle control signals by fine-tuning a LVLM on an instruction-tuning dataset. AgentDriver~\cite{mao2023language} converts driving situations into textual descriptions with human-like intelligence, then uses an LLM to reason and plan. Similarly, DriveVLM~\cite{sima2023drivelm} employs a LVLM to output planning trajectories through a Chain-of-Thought (CoT) reasoning process. 
However, due to being trained on vast amounts of general data, LVLMs often lack the specialized driving knowledge necessary for accuracy and reliability in driving. Therefore, a dataset that covers specific and comprehensive driving knowledge is crucial for both evaluating and enhancing LVLMs for autonomous driving.
%To address this, we present the first vision-language dataset focused specifically on explicit driving knowledge, which includes detailed questions on traffic regulations and driving techniques, significantly improving LVLM's understanding of driving complexities. Furthermore, our dataset allows for accurate comparison and evaluation of LVLMs in terms of driving knowledge, ultimately improving the accuracy and reliability of the models in practical applications.
\begin{figure*}[ht]
  \includegraphics[width=\linewidth]{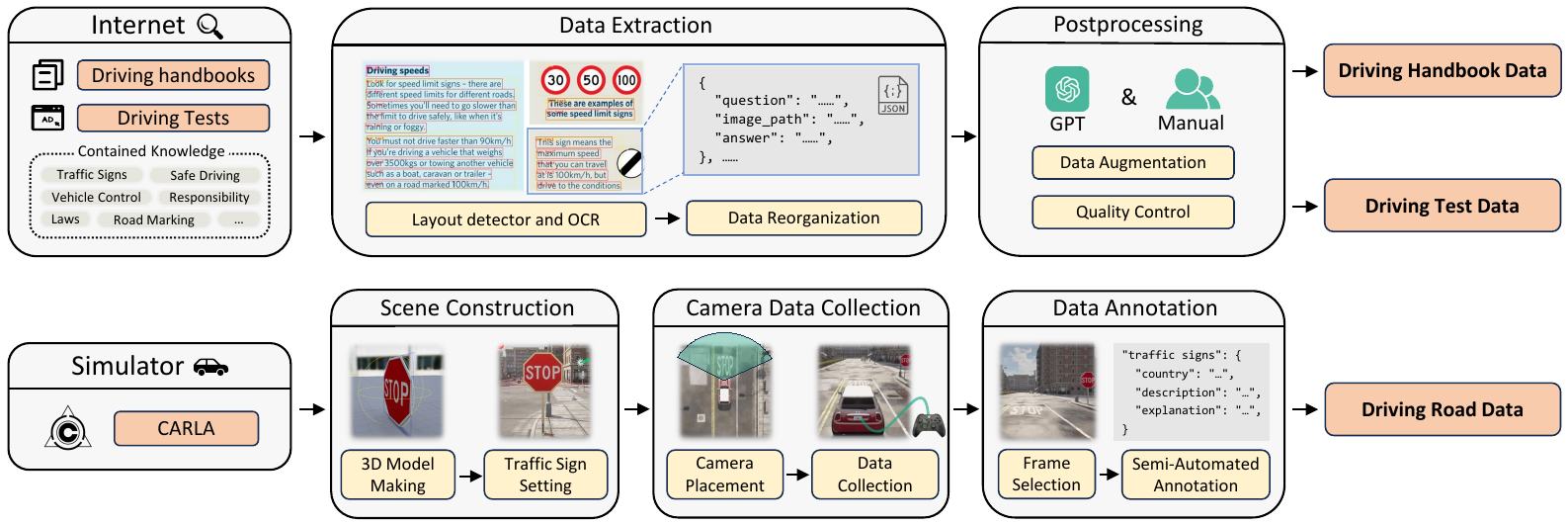}
  \caption{\textbf{Data construction pipeline of IDKB dataset.} For Driving Handbook and Driving Test Data, we collect comprehensive driving knowledge resources from internet, followed by data extraction and postprocessing to obtain the final data. For Driving Road Data, we utilize CARLA to generate simulated road scenarios focused on traffic sign comprehension.}
\label{fig:pipeline}
\end{figure*}

\subsection{Vision-Language Driving Datasets}

With the rise of Large Vision-Language Models (LVLMs), numerous vision-language datasets for autonomous driving have been developed for better understanding driving scenes.
The pioneering works BDD-X~\cite{kim2018textual} and BDD-OIA~\cite{xu2020explainable} annotated video datasets with textual descriptions and explanations of ego car actions. Many subsequent multimodal datasets have relabeled existing self-driving datasets. Talk2Car~\cite{deruyttere2019talk2car} adds free-form, high-quality natural language commands to the nuScenes dataset. NuScenes-QA~\cite{qian2024nuscenes} creates 460,000 question-answer pairs based on 3D object relationships to evaluate models' understanding and reasoning abilities. DriveLM~\cite{sima2023drivelm} constructs perception, prediction, and planning question-answer pairs in a graph structure to simulate human reasoning, thus enhancing end-to-end autonomous driving systems. VLAAD~\cite{park2024vlaad} utilizes GPT-4 to generate question-answer pairs from BDD-X~\cite{kim2018textual}, producing an instruction-following dataset that features complex reasoning, detailed descriptions, and conversation. However, these datasets rely heavily on existing datasets and often lack complex scenarios. To address this, some datasets are collected and annotated from scratch. DRAMA~\cite{malla2023drama} identifies critical objects in traffic scenarios and provides corresponding linguistic descriptions of driving risks. DriveVLM~\cite{tian2024drivevlm} presents SUP-AD, a scene understanding and planning dataset with annotations on challenging and long-tail scenarios. CODA-LM~\cite{li2024automated} is a large-scale multimodal self-driving dataset focusing on road corner cases.

However, these works lack explicit knowledge of traffic regulations, rules, and driving techniques, limiting LVLMs in their ability to develop a comprehensive and abstract understanding of driving knowledge.
%While their datasets may provide inplicit driving knowledge through scene understand and decision-making processes, this indirect approach limits a comprehensive and abstract understanding of driving knowledge for LVLMs. 
In contrast, our dataset mirrors the human process of acquiring driving knowledge by collecting detailed annotations from driving handbooks and test questions, while also integrating theoretical learning with practical application through simulated road scenarios in CARLA. By explicitly presenting this knowledge, our approach closely aligns with human learning styles, enabling LVLMs to acquire and integrate driving knowledge more efficiently and reliably, ultimately enhancing their performance in driving-related tasks.

%-------------------------------------------------------------------------
\section{Methods}
\label{sec:methods}

\textbf{I}ntelligent \textbf{D}riving \textbf{K}nowledge \textbf{B}ase (IDKB) is structured as a driving knowledge resource, mirroring the process individuals follow to acquire expertise when obtaining a driver's license. This process typically involves studying driving handbooks, taking theory tests, and practicing on the road. In this section, we introduce the data construction pipeline\ref{subsec: data_construction} and present the  statistics and characteristics of our dataset.

\subsection{Data Construction}
\label{subsec: data_construction}

\subsubsection{Driving Handbook Data} 
\begin{figure*}[ht]
  \includegraphics[width=\linewidth]{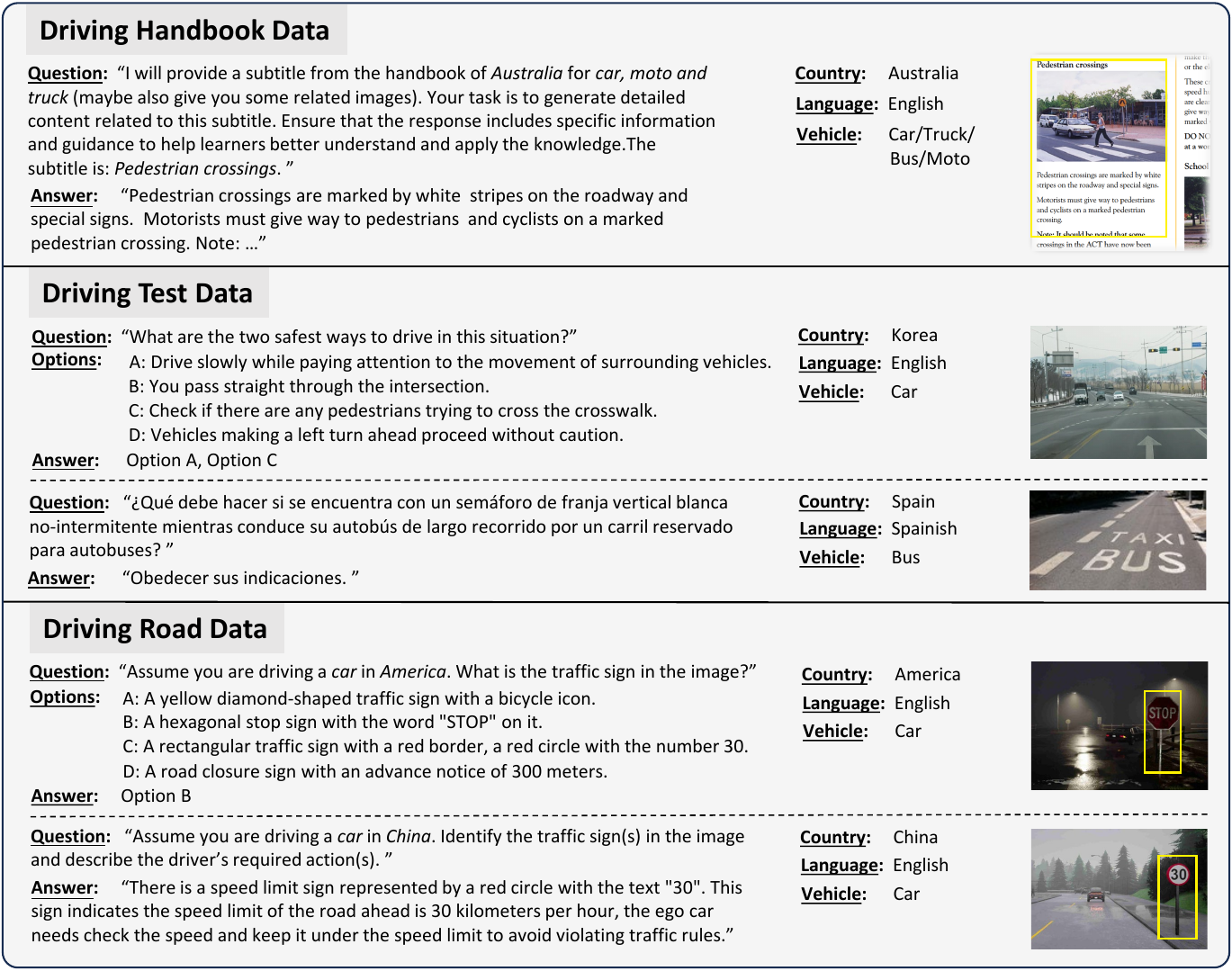}
  \caption{\textbf{Annotated examples of three data sources} -- Driving Handbook Data, Driving Test Data, and Driving Road Data.}
\label{fig:data_example}
\end{figure*}

Driving handbooks are highly structured and comprehensive resources, covering laws, regulations, techniques, safety, and more. They serve as the foundational step in learning to drive. By studying these handbooks, an intelligent system can develop a thorough and well-rounded understanding of the driving domain. We collected 206 documents, including traffic laws and driving handbooks, totalling 23,847 pages from 15 different countries via the Internet. As shown in Fig.~\ref{fig:pipeline}, we collect and organize unordered data from these documents in systematic way. We first employ layout detector and Optical Character Recognition (OCR) technology to extract data blocks and the text within these blocks. Subsequently, we developed an algorithm to cluster and sequence the data blocks in a way that aligns with human readability. Finally, we filter out duplicate and irrelevant data. An example is presented in Fig.~\ref{fig:data_example}. 
% In total, we extracted 7,733,124 words from the aforementioned documents.

\subsubsection{Driving Test Data}
Driving test data represent an alternative format for the knowledge covered in driving handbooks. While the handbooks offer a structured and comprehensive overview, the test questions reorganize this knowledge into multiple-choice and short-answer formats. This approach allows an intelligent system to both reinforce what it has learned and assess its understanding in a more interactive manner. Engaging with these questions ensures that the system has thoroughly internalized the handbook content, making this process an essential part of learning. To construct this part of data, we extensively collected questions from driving tests of 15 different countries. As depicted in Fig.~\ref{fig:pipeline}, the data collection process is similar with that of driving handbook data. We extract relevant information from various driving tests, and then reorganize these metadata into standard question-answer formats. Most entries we collected are multiple-choice questions, with one or more correct answers, while a smaller portion belongs to open QA questions. Two annotated examples are presented in Fig.~\ref{fig:data_example}. 
% Overall, we obtained a substantial amount of unaugmented driving test data, containing 50k+ well-annotated question-answer pairs from over 17 countries and regions. 

\textbf{Data Augmentation. }
To enhance the diversity of data and expand the scale of the dataset, we employed GPT-4o model to augment Driving Test Data. To ensure the enhanced data is well-structured, we divide the item into the question stem, options, and explanation sections. Each section undergoes incremental enhancement three times using GPT-4o. After completing the enhancements, the sections are combined to form the enhanced question-answer pair. To assure quality and avoid duplication, we require GPT-4o to output three distinct enhanced versions of the data in a single response, following a specific format. Invalid data is identified and removed through format checking and manual screening afterwards. 
In addition, for each data entry containing images, we generate textual descriptions of the images using GPT-4o and incorporate these descriptions into the dataset for further application.

\textbf{Data Quality Control. } 
To ensure high data quality, we implemented a two-step verification process at the end of the collection pipeline. Initially, an automated program filters out obvious low-quality data, such as images with extremely low resolution or very short text. After this automated removal, we conduct a manual review to further refine and ensure the quality of the remaining data.

\begin{table*}[ht]
\centering
\caption{\textbf{Comparison between our dataset and existing vision-language autonomous driving datasets.} ``QA'' means ``Question and Answer'' and ``MCQ'' indicates ``Multiple Choice Question''.}
\resizebox{\linewidth}{!}{
\begin{tabular}{l|cc|cc|ccc|cccc|c}
\toprule
\multirow{2}{*}{Dataset} &
  \multicolumn{2}{c|}{Data Type} &
  \multicolumn{2}{c|}{Data Source} &
  \multicolumn{3}{c|}{Data Domain} &
  \multicolumn{4}{c|}{Knowledge Domain} &
  \multirow{2}{*}{Amount} \\ \cmidrule{2-12}
 &
  QA &
  MCQ &
  Real &
  Synthetic &
  Country &
  Language &
  Vehicle Type &
  Laws \& Regulations &
  Signs \& Signals &
  Driving Techniques &
  \multicolumn{1}{l|}{Defensive Driving} &
   \\ \midrule
 BDD-X~\cite{kim2018textual} & \textcolor{red}{\xmark} & \textcolor{red}{\xmark} & \textcolor{green}{\cmark} & \textcolor{red}{\xmark} & US & EN & Car & \textcolor{red}{\xmark} & \textcolor{green}{\cmark} & \textcolor{green}{\cmark} & \textcolor{red}{\xmark} & 26K \\
  Talk2Car~\cite{deruyttere2019talk2car} & \textcolor{red}{\xmark} & \textcolor{red}{\xmark} & \textcolor{green}{\cmark} & \textcolor{red}{\xmark} & US, SG & EN & Car & \textcolor{red}{\xmark} & \textcolor{red}{\xmark} & \textcolor{red}{\xmark} & \textcolor{red}{\xmark} & 12K \\ 
  nuScenes-QA~\cite{qian2024nuscenes} & \textcolor{green}{\cmark} & \textcolor{red}{\xmark} & \textcolor{green}{\cmark} & \textcolor{red}{\xmark} & US, SG & EN & Car & \textcolor{red}{\xmark} & \textcolor{red}{\xmark} & \textcolor{red}{\xmark} & \textcolor{red}{\xmark} & 460K \\
  DriveLM~\cite{sima2023drivelm} & \textcolor{green}{\cmark} & \textcolor{red}{\xmark} & \textcolor{green}{\cmark} & \textcolor{green}{\cmark} & US, SG & EN & Car & \textcolor{red}{\xmark} & \textcolor{green}{\cmark} & \textcolor{green}{\cmark} & \textcolor{red}{\xmark} & 2M \\
  DRAMA~\cite{malla2023drama} & \textcolor{red}{\xmark} & \textcolor{red}{\xmark} & \textcolor{green}{\cmark} & \textcolor{red}{\xmark} & JP & EN & Car & \textcolor{red}{\xmark} & \textcolor{green}{\cmark} & \textcolor{green}{\cmark} & \textcolor{red}{\xmark} & 102K \\
  LangAuto CARLA~\cite{shao2024lmdrive} & \textcolor{red}{\xmark} & \textcolor{red}{\xmark} & \textcolor{red}{\xmark} & \textcolor{green}{\cmark} & US & EN & Car & \textcolor{red}{\xmark} & \textcolor{green}{\cmark} & \textcolor{green}{\cmark} & \textcolor{red}{\xmark} & 64K \\
  SUP-AD~\cite{tian2024drivevlm} & \textcolor{red}{\xmark} & \textcolor{red}{\xmark} & \textcolor{green}{\cmark} & \textcolor{red}{\xmark} & CN & EN & Car & \textcolor{red}{\xmark} & \textcolor{green}{\cmark} & \textcolor{green}{\cmark} & \textcolor{red}{\xmark} & - \\
  VLAAD~\cite{park2024vlaad} & \textcolor{green}{\cmark} & \textcolor{red}{\xmark} & \textcolor{green}{\cmark} & \textcolor{red}{\xmark} & US & EN & Car & \textcolor{red}{\xmark} & \textcolor{green}{\cmark} & \textcolor{green}{\cmark} & \textcolor{red}{\xmark} & 64K \\
  CODA-LM~\cite{li2022coda} & \textcolor{green}{\cmark} & \textcolor{red}{\xmark} & \textcolor{green}{\cmark} & \textcolor{red}{\xmark} & DE,CN,SG & EN & Car & \textcolor{red}{\xmark} & \textcolor{green}{\cmark} & \textcolor{green}{\cmark} & \textcolor{red}{\xmark} & 10K \\ \midrule
 \textbf{IDKB} & \textcolor{green}{\cmark} & \textcolor{green}{\cmark} & \textcolor{green}{\cmark} & \textcolor{green}{\cmark} & 15 & 9 & 4 & \textcolor{green}{\cmark} & \textcolor{green}{\cmark} & \textcolor{green}{\cmark} & \textcolor{green}{\cmark} & 1M \\ \bottomrule
\end{tabular}}
\label{tab:datasets_comparison}
\end{table*}

\begin{figure*}[h]
  \includegraphics[width=\linewidth]{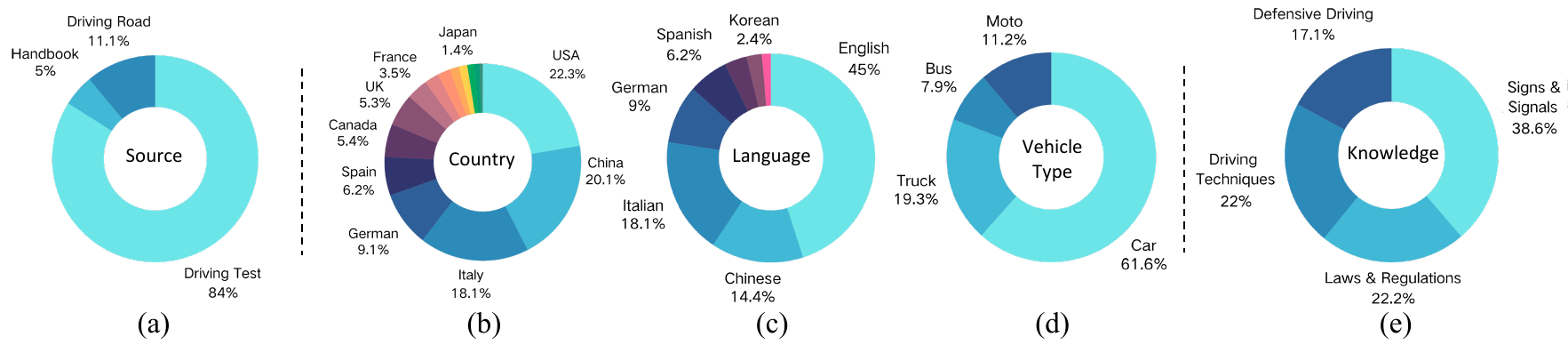}
  \caption{\textbf{Data distribution} in terms of data source, data domain and knowledge category.}
\label{fig:question_distribution}
\end{figure*}

\subsubsection{Driving Road Data}
After studying the driving handbooks and test questions, the intelligent system gains sufficient theoretical knowledge of driving. The next crucial step is to apply and reinforce this knowledge in real-world driving scenarios, which ensures that the system can effectively translate theoretical understanding into practical, real-world competence. However, existing datasets for real-world driving scenarios are often limited in scope, scale, and coverage of traffic signs, making it difficult to thoroughly understand road conditions. To address these limitations, we leverage the CARLA~\cite{dosovitskiy2017carla} simulator to generate a large, high-quality dataset that offers a more comprehensive understanding of traffic regulations at a low cost. 
Our approach begins with constructing custom simulated environments in the CARLA simulator to generate a batch of traffic sign understanding data. We further expand this dataset by extracting scenes containing traffic signs from the Bench2Drive (Jia et al., 2024) dataset, thereby creating additional annotated traffic sign data. As demonstrated in Fig.~\ref{fig:pipeline}, our driving road data collection process involves the following three steps.

\textbf{Scene Construction. }
In the scene construction stage, we first collect high-resolution traffic signs from different countries and create 3D models for each traffic sign using the CARLA-UE4 editor. Then we select two of CARLA’s large maps and set the traffic signs at appropriate locations within the CARLA simulator.

\textbf{Camera Data Collection. }
In the camera data collection stage, we generate an ego vehicle equipped with camera sensors to collect image data by driving the vehicle manually in the CARLA simulator. To ensure the authenticity of the simulation data, we randomly generate a large number of vehicles and pedestrians in the CARLA world and control them through the Autopilot mode and the WalkerAIController provided by the CARLA, respectively, to simulate the real road conditions. To ensure the diversity and richness of the data, we set different weather conditions, times of day, and road surface slipperiness to simulate real camera views and driving scenes. While collecting the camera sensor data, we also record the position, bounding box, rotation and other necessary information of each actor in the CARLA world for the next step. In total, we drive the ego vehicle in the CARLA simulator for about 20 hours and obtain approximately 400,000 frames of camera data.

\textbf{Data Annotation. }
After obtaining the image data and actor information, we automatically pick out the frames containing traffic signs based on the position, distance and direction of the sign relative to the ego vehicle. Then we manually build a dictionary to define the descriptions and explanations for each sign, and  attach the text annotations to each frame in an automated process. Finally, we obtained a total of 112,388 data samples, including both multiple-choice and question-and-answer formats. 

\textbf{Data Quality Control. }
Images that lack traffic signs or are obscured by adverse weather conditions or occlusions are considered as low-quality data. To maintain high data quality, we manually check and remove any images that do not meet the required standards.

\begin{table*}[ht]
\centering
\caption{\textbf{Quantitative results for multiple LVLMs on several tasks for driving knowledge understanding.} Results of proprietary LVLMs are overlaid in grey. The best results for each value of the proprietary LVLMs are highlighted in dark brown, while the best results for the open-source LVLMs are indicated in bold.}
\resizebox{\linewidth}{!}{
\begin{tabular}{l|c|cccc|cccc|c|ccc}
\toprule
\multirow{3}{*}{Model} &
  \multirow{3}{*}{IDKB Score} &
  \multicolumn{9}{c|}{Driving Test Data} &
  \multicolumn{3}{c}{Driving Road Data} \\ \cmidrule{3-14} 
 &
   &
  \multicolumn{4}{c|}{Multi-choice Question} &
  \multicolumn{4}{c|}{Question \& Answer} &
  \multirow{2}{*}{Test Data Score} &
  \multirow{2}{*}{Multi-choice Question} &
  \multirow{2}{*}{Question \& Answer} &
  \multirow{2}{*}{Road Data Score} \\ \cmidrule{3-10}
 &
   &
  Single Answer &
  Multiple Answer &
  Instruction Follow &
  \multicolumn{1}{c|}{Overall} &
  Rouge-1 &
  Rouge-L &
  SemScore &
  \multicolumn{1}{c|}{Overall} &
   &
   &
   \\ \midrule
   \rowcolor{gray!30}
GPT-4o \cite{achiam2023gpt}&
  \textbf{\textcolor[rgb]{0.4,0.26,0.15}{0.64}} &
  \textbf{\textcolor[rgb]{0.4,0.26,0.15}{0.54}} &
  0.40 &
  \textbf{\textcolor[rgb]{0.4,0.26,0.15}{0.99}} &
  \multicolumn{1}{c|}{\textbf{\textcolor[rgb]{0.4,0.26,0.15}{0.53}}} &
  \textbf{\textcolor[rgb]{0.4,0.26,0.15}{0.29}} &
  \textbf{\textcolor[rgb]{0.4,0.26,0.15}{0.27}} &
  \textbf{\textcolor[rgb]{0.4,0.26,0.15}{0.65}} &
  \multicolumn{1}{c|}{\textbf{\textcolor[rgb]{0.4,0.26,0.15}{0.54}}} & 
  \textbf{\textcolor[rgb]{0.4,0.26,0.15}{0.54}} &
  \textbf{\textcolor[rgb]{0.4,0.26,0.15}{0.87}} &
  0.59 &
  \textbf{\textcolor[rgb]{0.4,0.26,0.15}{0.73}} \\
  \rowcolor{gray!30}
Gemini-1.5-flash \cite{team2023gemini}&
  0.58 &
  0.40 &
  \textbf{\textcolor[rgb]{0.4,0.26,0.15}{0.44}} &
  \textbf{\textcolor[rgb]{0.4,0.26,0.15}{0.99}} &
  \multicolumn{1}{c|}{0.41} &
  0.19 &
  0.17 &
  0.55 &
  \multicolumn{1}{c|}{0.44} &
  0.43 &
  0.85 &
  \textbf{\textcolor[rgb]{0.4,0.26,0.15}{0.60}} &
  \textbf{\textcolor[rgb]{0.4,0.26,0.15}{0.73}} \\ \midrule 
VisualGLM-6B \cite{du2022glm}&
  0.29 &
  0.14 &
  0.01 &
  0.88 &
  \multicolumn{1}{c|}{0.12} &
  0.09 &
  0.07 &
  0.46 &
  \multicolumn{1}{c|}{0.35} &
  0.24 &
  0.18 &
  0.49 &
  0.34 \\
Qwen-VL-chat \cite{Qwen-VL}&
  0.37 &
  0.12 &
  0.01 &
  0.61 &
  \multicolumn{1}{c|}{0.11} &
  0.13 &
  0.11 &
  0.50 &
  \multicolumn{1}{c|}{0.39} &
  0.25 &
  0.47 &
  0.50 &
  0.49 \\
CogVLM \cite{wang2023cogvlm}&
  0.39 &
  \textbf{0.20} &
  0.03 &
  0.82 &
  \multicolumn{1}{c|}{\textbf{0.18}} &
  0.14 &
  0.12 &
  0.51 &
  \multicolumn{1}{c|}{0.40} &
  \textbf{0.29} &
  0.40 &
  0.58 &
  0.49 \\
BLIP2 \cite{li2023blip2bootstrappinglanguageimagepretraining}&
  0.27 &
  0.17 &
  0.26 &
  0.77 &
  \multicolumn{1}{c|}{\textbf{0.18}} &
  0.13 &
  0.12 &
  0.41 &
  \multicolumn{1}{c|}{0.32} &
  0.25 &
  0.17 &
  0.38 &
  0.28 \\
LLaVA-v1.5-7B \cite{liu2023llava}&
  0.36 &
  0.16 &
  0.32 &
  0.43 &
  \multicolumn{1}{c|}{0.17} &
  0.13 &
  0.12 &
  0.49 &
  \multicolumn{1}{c|}{0.38} &
  0.28 &
  0.33 &
  0.53 &
  0.43 \\
XComposer \cite{internlmxcomposer}&
  0.28 &
  0.14 &
  0.02 &
  \textbf{0.99} &
  \multicolumn{1}{c|}{0.13} &
  0.13 &
  0.12 &
  0.46 &
  \multicolumn{1}{c|}{0.36} &
  0.25 &
  0.24 &
  0.37 &
  0.31 \\
ShareGPT4V-7B \cite{chen2023sharegpt4v}&
  0.40 &
  0.17 &
  0.10 &
  \textbf{0.99} &
  \multicolumn{1}{c|}{0.16} &
  0.17 &
  0.16 &
  \textbf{0.52} &
  \multicolumn{1}{c|}{\textbf{0.41}} &
  \textbf{0.29} &
  0.44 &
  0.58 &
  0.51 \\
mPLUG-Owl2 \cite{ye2024mplug}&
  0.37 &
  0.18 &
  0.13 &
  0.59 &
  \multicolumn{1}{c|}{\textbf{0.18}} &
  0.14 &
  0.13 &
  0.51 &
  \multicolumn{1}{c|}{0.40} &
  \textbf{0.29} &
  0.39 &
  0.51 &
  0.45 \\
MiniCPM-Llama3-V2.5 \cite{yu2024rlaifv}&
  0.41 &
  0.11 &
  0.30 &
  0.71 &
  \multicolumn{1}{c|}{0.14} &
  0.05 &
  0.05 &
  0.43 &
  \multicolumn{1}{c|}{0.32} &
  0.23 &
  0.58 &
  \textbf{0.59} &
  0.59 \\
XComposer2 \cite{internlmxcomposer2}&
  \textbf{0.45} &
  0.11 &
  \textbf{0.35} &
  0.91 &
  \multicolumn{1}{c|}{0.14} &
  0.15 &
  0.14 &
  \textbf{0.52} &
  \multicolumn{1}{c|}{\textbf{0.41}} &
  0.28 &
  \textbf{0.62} &
  \textbf{0.59} &
  \textbf{0.61} \\
DeepSeek-VL-7B \cite{lu2024deepseekvl}&
  0.38 &
  0.09 &
  0.23 &
  0.47 &
  \multicolumn{1}{c|}{0.10} &
  0.15 &
  0.13 &
  \textbf{0.52} &
  \multicolumn{1}{c|}{\textbf{0.41}} &
  0.26 &
  0.45 &
  0.54 &
  0.50 \\
Yi-VL-6B \cite{ai2024yi}&
  0.28 &
  0.15 &
  0.27 &
  0.11 &
  \multicolumn{1}{c|}{0.16} &
  0.14 &
  0.13 &
  0.47 &
  \multicolumn{1}{c|}{0.37} &
  0.27 &
  0.30 &
  0.27 &
  0.29 \\
Monkey-chat \cite{li2023monkey}&
  0.39 &
  0.11 &
  0.01 &
  0.88 &
  \multicolumn{1}{c|}{0.10} &
  \textbf{0.18} &
  \textbf{0.17} &
  0.50 &
  \multicolumn{1}{c|}{0.40} &
  0.25 &
  0.46 &
  0.58 &
  0.52 \\ \bottomrule
\end{tabular}}
\label{tab:overall_eval}
\end{table*}

\subsection{Dataset Statistics}
In total, IDKB provides 1,016,956 data entries. As shown in Fig.~\ref{fig:question_distribution}(a), Driving Test Data constitutes the largest portion of the dataset (84.0\%), with CARLA and Driving Handbook Data accounting for 11.1\% and 5.0\%, respectively. Our dataset spans multiple countries and multiple languages. As presented in Fig.~\ref{fig:question_distribution}(b)(c), apart from English-speaking countries, we also include driving knowledge from China, Italy, Germany, and others. In terms of vehicle types, we categorized the data into four classes: Car (standard passenger vehicles including sedan, jeep, ...), Truck (large vehicles including minivan, commercial, LGV, ...), Bus (including minibus, trailer, coaches, ...), and Moto (Motobike, Motocycle). Fig.~\ref{fig:question_distribution}(d) presents the distribution of the vehicle types.

To better analyze the knowledge coverage of our dataset, we employed proprietary LVLMs to classify all the questions into four major categories according to their semantics, including Laws \& Regulations (22.2\%), Road Signs \& Signals (38.6\%), Driving Techniques (22.0\%), and Defensive Driving (17.1\%). More data details are provided in Supplementary.

\subsection{Data Characteristics}
As shown in Tab.~\ref{tab:datasets_comparison}, compared with existing vision-language autonomous driving datasets, IDKB possesses four main novel characteristics as follows.

\subsubsection{Diverse Data Type} 
Our dataset contains diverse data types, encompassing both Question and Answer (QA) and Multiple-choice Question (MCQ). Most existing datasets typically focus on a single question format. By including both QA and MCQ formats, IDKB enhances its utility for various applications, such as training models for open-ended and structured queries, thus providing a more comprehensive testing ground for autonomous driving models.

\subsubsection{Diverse Data Source}
Our dataset integrates both real-world and synthetic data sources to provide a comprehensive coverage of driving scenarios. We collected driving manuals and test questions from various countries across the internet, which form the basis of our real-world data. To complement this, we enriched our dataset with data from CARLA-simulated road scenes. Relying solely on  real-world road data has its limitations, as it may not cover all possible scenarios encountered in diverse driving environments. By incorporating CARLA simulations, we address this limitation and ensure that our dataset encompasses a wider range of scenarios. This combined approach allows the system to effectively translate theoretical knowledge from driving manuals and test questions into practical operational capabilities.

\subsubsection{Diverse Data Domain}
IDKB exhibits exceptional domain diversity, covering 15 different countries, 9 languages, and 4 types of vehicles. This extensive coverage makes the dataset particularly versatile, enabling its application to various regional contexts and linguistic environments. Most existing datasets are limited to a specific country or language, typically focusing on the US and English. In contrast, IDKB's global approach ensures that models trained on it are better equipped to handle international variations in driving conditions, regulations, and vehicle types, facilitating broader applicability in autonomous driving systems worldwide.

\begin{table*}[ht]
\centering
\caption{\textbf{Comparison of results for fine-tuned LVLMs.} ``MCQ'' refers to multi-choice question and ``QA'' denotes question-and-answer question,  ``Improvements'' represents the percentage increase in the three scores compared to the original value. Results of the proprietary LVLM, which has not been fine-tuned and is included for comparison, are overlaid in grey. The best results for the open-source LVLMs are indicated in bold.}
\resizebox{\linewidth}{!}{
\begin{tabular}{l|cc|cccc|cccc}
\toprule
Model &
  \multirow{2}{*}{IDKB Score} &
  \multirow{2}{*}{Improvement} &
  \multicolumn{4}{c|}{Driving Test Data} &
  \multicolumn{4}{c}{Driving Road Data} \\ \cmidrule{4-11} 
 &
   &
   &
  MCQ &
  QA &
  \multicolumn{1}{c|}{Test Data Score} &
  Improvement &
  MCQ &
  QA &
  \multicolumn{1}{c|}{Road Data Score} &
  Improvement \\ \midrule
  \rowcolor{gray!30}
GPT-4o \cite{achiam2023gpt} &
  0.64 &
  N/A &
  0.53 &
  0.54 &
  \multicolumn{1}{c|}{0.54} &
  N/A &
  0.87 &
  0.60 &
  \multicolumn{1}{c|}{0.74} &
  N/A \\ \midrule
Qwen-VL-chat \cite{Qwen-VL} &
  0.56 &
  51\% &
  0.40 &
   0.50 &
  \multicolumn{1}{c|}{0.45} &
  80\% &
  0.70 &
   0.64 &
  \multicolumn{1}{c|}{0.67} &
   37\% \\
MiniCPM-Llama3-V2.5 \cite{yu2024rlaifv} &
  0.62 &
  42\% &
  0.42 &
   0.49 &
  \multicolumn{1}{c|}{0.46} &
  \textbf{100\%} &
  \textbf{0.85} &
   0.68&
  \multicolumn{1}{c|}{\textbf{0.77}} &
   31\% \\
XComposer2\cite{internlmxcomposer2} &
  \textbf{0.64} &
  56\% &
  \textbf{0.51} &
  \textbf{0.52} &
  \multicolumn{1}{c|}{\textbf{0.52}} &
  86\% &
  0.81 &
  \textbf{0.71} &
  \multicolumn{1}{c|}{0.76} &
  25\% \\
DeepSeek-VL-7B \cite{lu2024deepseekvl} &
  0.60 &
  \textbf{58\%} &
  0.48 &
  0.51 &
  \multicolumn{1}{c|}{0.50} &
  92\% &
  0.68 &
  0.69 &
  \multicolumn{1}{c|}{0.69} &
  \textbf{38\%} \\ \bottomrule
\end{tabular}
}
\label{tab: finetune}
\end{table*}

\subsubsection{Diverse Knowledge Domain}
Our dataset also offers comprehensive coverage of knowledge domains relevant to autonomous driving. It includes detailed information on Traffic Laws and Regulations, Road Signs and Signals, Vehicle Control and Driving Techniques, and Defensive Driving strategies. While other datasets may focus on one or two knowledge areas, IDKB provides a holistic view of the driving environment. This broad knowledge diversity is crucial for developing models that need to understand and navigate complex driving scenarios, making IDKB an invaluable resource for advancing autonomous driving technologies.

%-------------------------------------------------------------------------
\section{Experiments}
\label{sec:exp}
In this section, we evaluate 15 Large Vision-Language Models (LVLMs), both open-source and closed-source, using our proposed dataset. We begin by introducing the selected LVLMs and the tasks they are required to perform. Then, we outline the evaluation methods used for each task. Next, we present a quantitative evaluation of each task.

\subsection{Experiment Setup}

\textbf{Selected LVLMs.} We test 15 representative LVLMs that differ in terms of parameters, open-source availability, and their vision encoders (CLIP ViT~\cite{radford2021learning}, EVA-CLIP-ViT~\cite{sun2023eva}, SAM~\cite{kirillov2023segment}, SigLIP~\cite{zhai2023sigmoid}) as well as their LLMs (QWen~\cite{bai2023qwen}, Vicuna~\cite{zheng2024judging}, Yi~\cite{young2024yi}, DeepSeek~\cite{deepseek-llm}, InternLM~\cite{cai2024internlm2}, LLaMA~\cite{touvron2023llama}, ChatGLM~\cite{glm2024chatglm}, FLAN-T5~\cite{chung2024scaling}). For a fair comparison, all LVLMs are used to infer questions from our dataset based on the same prompt. Further details on the selected LVLMs are provided in the Supplementary.

\noindent\textbf{Data Split.} We selected all of the driving handbook data, along with ninety percent of the driving test data and driving road data, to form the training set. The remaining ten percent of the driving test data and driving road data were used to create the test set. For more detailed statistics on the training and test sets, please refer to the Supplementary.

\noindent\textbf{Tasks.} We evaluate the LVLM's performance using two data sources: driving test data and driving road data, each containing multiple-choice questions (MCQ) and question-and-answer (QA) tasks. In the MCQ tasks, the LVLM must select the correct answer(s) from the provided options, while in the QA tasks, it is required to generate the most relevant response to a given question.

\subsection{Evaluation Details}

For MCQ tasks, we use regular expressions to extract options from the LVLM outputs and compare them with correct answers, measuring accuracy as the metric. Partially correct answers are not accepted in multi-answer questions. Rule-based extraction is challenging due to the free-form nature of LVLM outputs. To address this, we introduce a instruction-following test where the output must include the string ``Option [A to F]'' as prompted.

For QA tasks, we use ROUGE~\cite{lin2004rouge} and SEMScore~\cite{aynetdinov2024semscoreautomatedevaluationinstructiontuned} to measure similarity between LVLM outputs and the reference answers. ROUGE evaluates N-gram overlap, while SEMScore assesses semantic similarity using sentence embeddings.

We calculate average scores for both MCQ and QA metrics across data sources, creating Test Data and Road Data Scores. The mean of these two scores is the IDKB Score, reflecting the LVLM's overall mastery of driving knowledge. More details about metrics are provided in Supplementary.

\subsection{Main Results}
In this subsection, we analyze various LVLMs on our test set, focusing on overall performance, distinctions between driving test and road data, handling of single versus multiple answers, comparison between proprietary and open-source LVLMs, and adherence to instructions. We also highlight the impact of fine-tuning with our dataset, showcasing key improvements in model performance. More details and analysis will be provided in Supplementary. 

\begin{table*}[ht]
\centering
\caption{\textbf{Comparison of planning results on the nuScenes validation dataset using UniAD~\cite{hu2023planning} metrics.}}
\begin{tabular}{l|cccc|cccc}
\toprule
\multirow{2}{*}{Setup} & \multicolumn{4}{c|}{L2 (m)} & \multicolumn{4}{c}{Collision (\%)} \\ \cmidrule{2-9} 
                       & 1s    & 2s    & 3s   & Avg. & 1s      & 2s     & 3s     & Avg.   \\ \midrule
nuScenes finetune only (w/o IDKB)
     & 0.50  & 0.98  & 1.91 & 1.13 & 0.13    & 0.32   & 1.38   & 0.61   \\
nuScenes + IDKB fine-tuning       & \textbf{0.23}  & \textbf{0.68}  & \textbf{1.40} & \textbf{0.77} & \textbf{0.07}    & \textbf{0.15}   & \textbf{0.90}   & \textbf{0.37}   \\ \bottomrule
\end{tabular}
\label{tab:compare_planning}
\end{table*}

\textbf{Overall Evaluation} 
In Tab.~\ref{tab:overall_eval}, we present the performance results of various VLMs on our test set. GPT-4o achieved the highest overall performance with an IDKB score of 0.64. Among the open-source models, XComposer2 stood out with the best performance, achieving an IDKB score of 0.45. Most open-source models fell within an IDKB score range from 0.35 to 0.4. However, BLIP2, XComposer, Yi-VL and VisualGLM underperformed, with IDKB scores of 0.27, 0.28, 0.28 and 0.29, respectively. Overall, the evaluated LVLMs generally did not demonstrate strong driving domain knowledge, highlighting the need for high-quality, structured and diverse driving knowledge data for effective applications in autonomous driving.

\textbf{Driving Test Data vs. Driving Road Data }
LVLMs generally perform better on driving road data than on driving test data. Most open-source LVLMs scored around 0.25 on Test Data Score, while the average Road Data Score was 0.44. A similar trend was observed in the two proprietary LVLMs. This suggests that while many LVLMs have a basic understanding of traffic signs, they lack a deeper comprehension of traffic laws, regulations, and driving skills—areas more thoroughly assessed in driving test data.

\textbf{Single Answer vs. Multiple Answer}
The performance of LVLMs on multi-answer questions showed clear polarization. About half of the LVLMs outperformed on multi-answer questions compared to single-answer questions. Conversely, the other half performed significantly worse on multi-answer questions than on single-answer ones. Notably, models such as VisualGLM-6B, Qwen-VL, CogVLM, XComposer, and Monkey-chat struggled particularly with multi-answer questions, failing to provide accurate responses. This disparity in performance suggests that the complexity of driving knowledge required for multi-answer questions presents a substantial challenge for certain VLMs, underscoring the variability in their capabilities.

\textbf{Instruction Follow Ability}
Overall, proprietary models excel in adherence to instructions, with accuracy of 0.99, indicating minimal deviation from prescribed answer templates. In contrast, among open-source models, only VisualGLM-6B, XComposer, ShareGPT4V-6B, XComposer2, and Monkey-chat demonstrate a relatively high level of instruction-following capability. The remaining open-source models are prone to producing non-compliant text, with Yi-VL-6B exhibiting the most significant issue—only 11\% of its outputs align with the input requirements. Some LVLMs may struggle with consistency and accuracy in adhering to specified instructions, underscoring the need for additional fine-tuning to enhance their ability to comply with instructions in practical applications.

\textbf{Proprietary vs. Open-Source}
Proprietary LVLMs generally outperform open-source LVLMs in evaluation results, likely due to their larger number of parameters and more extensive knowledge base. This advantage is more pronounced with driving test data, which requires external knowledge of laws and rules—areas where proprietary models excel. In contrast, when dealing with driving road data, which emphasizes traffic sign recognition, some open-source LVLMs like XComposer2 can achieve performance levels comparable to those of proprietary models.

% \textbf{Hard cases for VLMs}

\textbf{Significant Improvement through Fine-Tuning}
To better evaluate the impact of structured driving knowledge data on model performance in this domain, we fine-tuned four representative LVLMs, each with a different visual encoder or LLM. As shown in Table \ref{tab: finetune}, the fine-tuned LVLMs achieved IDKB scores comparable to, or even matching, those of proprietary models with significantly larger parameters.
The improvement in the Test Data Score is particularly striking, with MiniCPM-Llama3-V2.5's performance doubling in this metric. This suggests that our data equips LVLMs with valuable expertise in driving laws, rules, techniques, and handling special situations. 
Additionally, the model's ability to interpret traffic signs, as indicated by the Test Road Score, showed improvement over the original, suggesting an enhanced understanding of traffic regulations in road scenarios. 
These results underscore the importance of our dataset in enhancing LVLM competence within the driving domain. By equipping models with structured and diverse driving knowledge, our dataset plays a crucial role in strengthening LVLMs' expertise, ultimately contributing to the development of safer and more reliable autonomous driving systems.

%-------------------------------------------------------------------------

\section{Benefits of Driving Knowledge for Downstream Autonomous Driving Tasks}
\label{sec:Benefits_of_Driving_Knowledge}
In this section, we showcase the application of the Qwen-VL-chat model, fine-tuned on the IDKB dataset, for the planning task using the nuScenes~\cite{caesar2020nuscenes} dataset. This demonstrates how integrating driving knowledge data can significantly enhance performance in downstream tasks.

\subsection{Setup}
Based on Agent-Driver~\cite{agentdriver}, we generated fine-tuning data for nuScenes trajectory planning and fine-tuned Qwen-VL-chat for this task. To demonstrate the value of IDKB data for autonomous driving, we also fine-tuned another Qwen-VL-chat model using a combination of the nuScenes planning data and our IDKB data.

\begin{figure*}[ht]
 \centering
  \includegraphics[width=0.8\textwidth]{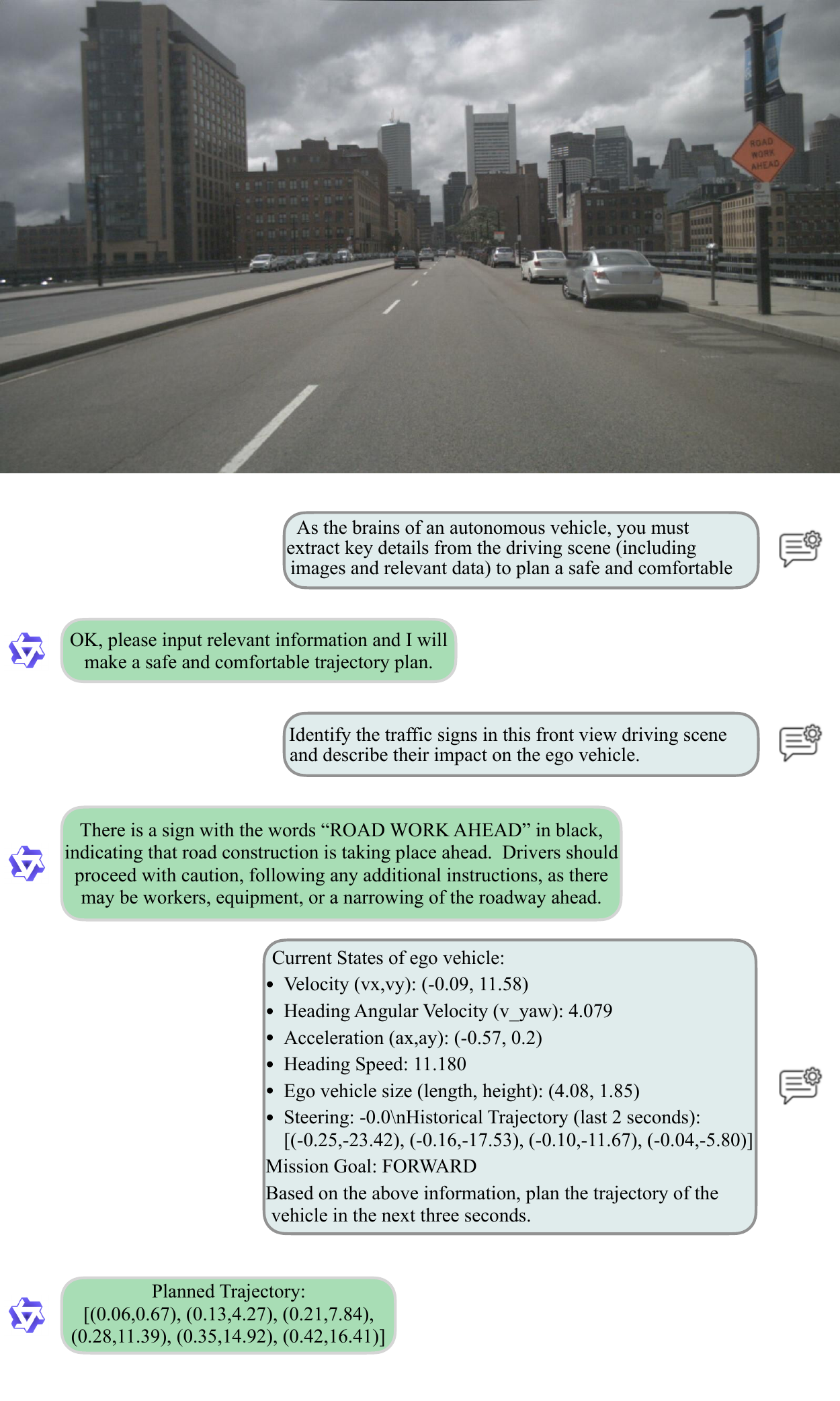}
  \caption{\textbf{Visualization of LVLM’s inference process.} Qwen-VL-chat, fine-tuned on both IDKB and nuScenes, identifies the traffic sign ahead and recommends a driving decision to slow down.}
\label{fig:vis_planning}
\end{figure*}

For inference, as shown in Fig.~\ref{fig:vis_planning}, we follow a prompt method similar to DriveVLM~\cite{DriveVLM}. The LVLM first identifies the traffic signs on the road and then receives key information to predict the trajectory for the next three seconds.

\afterpage{\clearpage}

\subsection{Results and analysis}
The nuScenes planning task relies on imitation learning from human-generated planning data, while the IDKB dataset is designed to align with the human model of driving knowledge acquisition. This alignment allows the LVLM fine-tuned with IDKB to gain a deeper understanding of driving knowledge, including traffic laws, regulations, and driving skills, which in turn leads to more rational and safer route planning.

Our experimental results support this hypothesis. As shown in the table, the model fine-tuned with IDKB data demonstrates superior performance, with a 32\% reduction in average L2 distance and a 65\% decrease in collision metrics, indicating safer and more rational planning.

Additionally, as illustrated in Fig.~\ref{fig:vis_planning}, the model fine-tuned with both nuScenes and IDKB data successfully identifies the 'ROAD WORK AHEAD' sign, understands the need to slow down, and drive with caution. The planned trajectory confirms this, with the decreasing offset between consecutive frames reflecting deceleration.

%-------------------------------------------------------------------------
\section{Conclusions}
\label{sec:conclusions}
In this paper, we introduced IDKB, a pioneering large-scale dataset designed to bridge the gap in domain-specific driving knowledge within LVLMs. IDKB includes over 1 million entries on driving regulations, scenarios, and practices from 15 countries and 9 languages. 
We evaluated 15 LVLMs using IDKB to assess their performance in the context of autonomous driving and provided extensive analysis. Additionally, we fine-tuned several popular models, achieving significant improvements in their performance, which further highlights the importance of our dataset.

%-------------------------------------------------------------------------

%%%%%%%%% REFERENCES
{
\small
\bibliographystyle{ieeenat_fullname}
\bibliography{main}

\begin{thebibliography}{52}
\providecommand{\natexlab}[1]{#1}
\providecommand{\url}[1]{\texttt{#1}}
\expandafter\ifx\csname urlstyle\endcsname\relax
  \providecommand{\doi}[1]{doi: #1}\else
  \providecommand{\doi}{doi: \begingroup \urlstyle{rm}\Url}\fi

\bibitem[Achiam et~al.(2023)Achiam, Adler, Agarwal, Ahmad, Akkaya, Aleman, Almeida, Altenschmidt, Altman, Anadkat, et~al.]{achiam2023gpt}
Josh Achiam, Steven Adler, Sandhini Agarwal, Lama Ahmad, Ilge Akkaya, Florencia~Leoni Aleman, Diogo Almeida, Janko Altenschmidt, Sam Altman, Shyamal Anadkat, et~al.
\newblock Gpt-4 technical report.
\newblock \emph{arXiv preprint arXiv:2303.08774}, 2023.

\bibitem[AI et~al.(2024)AI, :, Young, Chen, Li, Huang, Zhang, Zhang, Li, Zhu, Chen, Chang, Yu, Liu, Liu, Yue, Yang, Yang, Yu, Xie, Huang, Hu, Ren, Niu, Nie, Xu, Liu, Wang, Cai, Gu, Liu, and Dai]{ai2024yi}
01. AI, :, Alex Young, Bei Chen, Chao Li, Chengen Huang, Ge Zhang, Guanwei Zhang, Heng Li, Jiangcheng Zhu, Jianqun Chen, Jing Chang, Kaidong Yu, Peng Liu, Qiang Liu, Shawn Yue, Senbin Yang, Shiming Yang, Tao Yu, Wen Xie, Wenhao Huang, Xiaohui Hu, Xiaoyi Ren, Xinyao Niu, Pengcheng Nie, Yuchi Xu, Yudong Liu, Yue Wang, Yuxuan Cai, Zhenyu Gu, Zhiyuan Liu, and Zonghong Dai.
\newblock Yi: Open foundation models by 01.ai, 2024.

\bibitem[Aynetdinov and Akbik(2024)]{aynetdinov2024semscoreautomatedevaluationinstructiontuned}
Ansar Aynetdinov and Alan Akbik.
\newblock Semscore: Automated evaluation of instruction-tuned llms based on semantic textual similarity, 2024.

\bibitem[Bai et~al.(2023{\natexlab{a}})Bai, Bai, Yang, Wang, Tan, Wang, Lin, Zhou, and Zhou]{Qwen-VL}
Jinze Bai, Shuai Bai, Shusheng Yang, Shijie Wang, Sinan Tan, Peng Wang, Junyang Lin, Chang Zhou, and Jingren Zhou.
\newblock Qwen-vl: A versatile vision-language model for understanding, localization, text reading, and beyond.
\newblock \emph{arXiv preprint arXiv:2308.12966}, 2023{\natexlab{a}}.

\bibitem[Bai et~al.(2023{\natexlab{b}})Bai, Bai, Yang, Wang, Tan, Wang, Lin, Zhou, and Zhou]{bai2023qwen}
Jinze Bai, Shuai Bai, Shusheng Yang, Shijie Wang, Sinan Tan, Peng Wang, Junyang Lin, Chang Zhou, and Jingren Zhou.
\newblock Qwen-vl: A frontier large vision-language model with versatile abilities.
\newblock \emph{arXiv preprint arXiv:2308.12966}, 2023{\natexlab{b}}.

\bibitem[Caesar et~al.(2020)Caesar, Bankiti, Lang, Vora, Liong, Xu, Krishnan, Pan, Baldan, and Beijbom]{caesar2020nuscenes}
Holger Caesar, Varun Bankiti, Alex~H Lang, Sourabh Vora, Venice~Erin Liong, Qiang Xu, Anush Krishnan, Yu Pan, Giancarlo Baldan, and Oscar Beijbom.
\newblock nuscenes: A multimodal dataset for autonomous driving.
\newblock In \emph{Proceedings of the IEEE/CVF conference on computer vision and pattern recognition}, pages 11621--11631, 2020.

\bibitem[Cai and et~al.(2024)]{cai2024internlm2}
Zheng Cai and Maosong~Cao et al.
\newblock Internlm2 technical report, 2024.

\bibitem[Chen et~al.(2023)Chen, Li, Dong, Zhang, He, Wang, Zhao, and Lin]{chen2023sharegpt4v}
Lin Chen, Jisong Li, Xiaoyi Dong, Pan Zhang, Conghui He, Jiaqi Wang, Feng Zhao, and Dahua Lin.
\newblock Sharegpt4v: Improving large multi-modal models with better captions.
\newblock \emph{arXiv preprint arXiv:2311.12793}, 2023.

\bibitem[Chung et~al.(2024)Chung, Hou, Longpre, Zoph, Tay, Fedus, Li, Wang, Dehghani, Brahma, et~al.]{chung2024scaling}
Hyung~Won Chung, Le Hou, Shayne Longpre, Barret Zoph, Yi Tay, William Fedus, Yunxuan Li, Xuezhi Wang, Mostafa Dehghani, Siddhartha Brahma, et~al.
\newblock Scaling instruction-finetuned language models.
\newblock \emph{Journal of Machine Learning Research}, 25\penalty0 (70):\penalty0 1--53, 2024.

\bibitem[DeepSeek-AI(2024)]{deepseek-llm}
DeepSeek-AI.
\newblock Deepseek llm: Scaling open-source language models with longtermism.
\newblock \emph{arXiv preprint arXiv:2401.02954}, 2024.

\bibitem[Deruyttere et~al.(2019)Deruyttere, Vandenhende, Grujicic, Van~Gool, and Moens]{deruyttere2019talk2car}
Thierry Deruyttere, Simon Vandenhende, Dusan Grujicic, Luc Van~Gool, and Marie-Francine Moens.
\newblock Talk2car: Taking control of your self-driving car.
\newblock \emph{arXiv preprint arXiv:1909.10838}, 2019.

\bibitem[Dong et~al.(2024)Dong, Zhang, Zang, Cao, Wang, Ouyang, Wei, Zhang, Duan, Cao, Zhang, Li, Yan, Gao, Zhang, Li, Li, Chen, He, Zhang, Qiao, Lin, and Wang]{internlmxcomposer2}
Xiaoyi Dong, Pan Zhang, Yuhang Zang, Yuhang Cao, Bin Wang, Linke Ouyang, Xilin Wei, Songyang Zhang, Haodong Duan, Maosong Cao, Wenwei Zhang, Yining Li, Hang Yan, Yang Gao, Xinyue Zhang, Wei Li, Jingwen Li, Kai Chen, Conghui He, Xingcheng Zhang, Yu Qiao, Dahua Lin, and Jiaqi Wang.
\newblock Internlm-xcomposer2: Mastering free-form text-image composition and comprehension in vision-language large model.
\newblock \emph{arXiv preprint arXiv:2401.16420}, 2024.

\bibitem[Dosovitskiy et~al.(2017)Dosovitskiy, Ros, Codevilla, Lopez, and Koltun]{dosovitskiy2017carla}
Alexey Dosovitskiy, German Ros, Felipe Codevilla, Antonio Lopez, and Vladlen Koltun.
\newblock Carla: An open urban driving simulator.
\newblock In \emph{Conference on robot learning}, pages 1--16. PMLR, 2017.

\bibitem[Du et~al.(2022)Du, Qian, Liu, Ding, Qiu, Yang, and Tang]{du2022glm}
Zhengxiao Du, Yujie Qian, Xiao Liu, Ming Ding, Jiezhong Qiu, Zhilin Yang, and Jie Tang.
\newblock Glm: General language model pretraining with autoregressive blank infilling.
\newblock In \emph{Proceedings of the 60th Annual Meeting of the Association for Computational Linguistics (Volume 1: Long Papers)}, pages 320--335, 2022.

\bibitem[Ester et~al.(1996)Ester, Kriegel, Sander, and Xu]{DBSCAN}
Martin Ester, Hans-Peter Kriegel, J\"{o}rg Sander, and Xiaowei Xu.
\newblock A density-based algorithm for discovering clusters in large spatial databases with noise.
\newblock In \emph{Proceedings of the Second International Conference on Knowledge Discovery and Data Mining}, page 226–231. AAAI Press, 1996.

\bibitem[et~al.(2024)]{glm2024chatglm}
Team~GLM et al.
\newblock Chatglm: A family of large language models from glm-130b to glm-4 all tools, 2024.

\bibitem[Hu et~al.(2021)Hu, Shen, Wallis, Allen-Zhu, Li, Wang, Wang, and Chen]{hu2021lora}
Edward~J Hu, Yelong Shen, Phillip Wallis, Zeyuan Allen-Zhu, Yuanzhi Li, Shean Wang, Lu Wang, and Weizhu Chen.
\newblock Lora: Low-rank adaptation of large language models.
\newblock \emph{arXiv preprint arXiv:2106.09685}, 2021.

\bibitem[Hu et~al.(2023)Hu, Yang, Chen, Li, Sima, Zhu, Chai, Du, Lin, Wang, et~al.]{hu2023planning}
Yihan Hu, Jiazhi Yang, Li Chen, Keyu Li, Chonghao Sima, Xizhou Zhu, Siqi Chai, Senyao Du, Tianwei Lin, Wenhai Wang, et~al.
\newblock Planning-oriented autonomous driving.
\newblock In \emph{Proceedings of the IEEE/CVF Conference on Computer Vision and Pattern Recognition}, pages 17853--17862, 2023.

\bibitem[Huang et~al.(2022)Huang, Lv, Cui, Lu, and Wei]{huang2022layoutlmv3}
Yupan Huang, Tengchao Lv, Lei Cui, Yutong Lu, and Furu Wei.
\newblock Layoutlmv3: Pre-training for document ai with unified text and image masking.
\newblock In \emph{Proceedings of the 30th ACM International Conference on Multimedia}, 2022.

\bibitem[Kim et~al.(2018)Kim, Rohrbach, Darrell, Canny, and Akata]{kim2018textual}
Jinkyu Kim, Anna Rohrbach, Trevor Darrell, John Canny, and Zeynep Akata.
\newblock Textual explanations for self-driving vehicles.
\newblock In \emph{Proceedings of the European conference on computer vision (ECCV)}, pages 563--578, 2018.

\bibitem[Kirillov et~al.(2023)Kirillov, Mintun, Ravi, Mao, Rolland, Gustafson, Xiao, Whitehead, Berg, Lo, et~al.]{kirillov2023segment}
Alexander Kirillov, Eric Mintun, Nikhila Ravi, Hanzi Mao, Chloe Rolland, Laura Gustafson, Tete Xiao, Spencer Whitehead, Alexander~C Berg, Wan-Yen Lo, et~al.
\newblock Segment anything.
\newblock In \emph{Proceedings of the IEEE/CVF International Conference on Computer Vision}, pages 4015--4026, 2023.

\bibitem[Li et~al.(2023)Li, Li, Savarese, and Hoi]{li2023blip2bootstrappinglanguageimagepretraining}
Junnan Li, Dongxu Li, Silvio Savarese, and Steven Hoi.
\newblock Blip-2: Bootstrapping language-image pre-training with frozen image encoders and large language models, 2023.

\bibitem[Li et~al.(2022)Li, Chen, Wang, Hong, Ye, Han, Chen, Zhang, Xu, Yeung, et~al.]{li2022coda}
Kaican Li, Kai Chen, Haoyu Wang, Lanqing Hong, Chaoqiang Ye, Jianhua Han, Yukuai Chen, Wei Zhang, Chunjing Xu, Dit-Yan Yeung, et~al.
\newblock Coda: A real-world road corner case dataset for object detection in autonomous driving.
\newblock In \emph{European Conference on Computer Vision}, pages 406--423. Springer, 2022.

\bibitem[Li et~al.(2024{\natexlab{a}})Li, Zhang, Chen, Liu, Li, Gao, Hong, Tian, Zhao, Li, et~al.]{li2024automated}
Yanze Li, Wenhua Zhang, Kai Chen, Yanxin Liu, Pengxiang Li, Ruiyuan Gao, Lanqing Hong, Meng Tian, Xinhai Zhao, Zhenguo Li, et~al.
\newblock Automated evaluation of large vision-language models on self-driving corner cases.
\newblock \emph{arXiv preprint arXiv:2404.10595}, 2024{\natexlab{a}}.

\bibitem[Li et~al.(2024{\natexlab{b}})Li, Yang, Liu, Ma, Zhang, Yang, Sun, Liu, and Bai]{li2023monkey}
Zhang Li, Biao Yang, Qiang Liu, Zhiyin Ma, Shuo Zhang, Jingxu Yang, Yabo Sun, Yuliang Liu, and Xiang Bai.
\newblock Monkey: Image resolution and text label are important things for large multi-modal models.
\newblock In \emph{proceedings of the IEEE/CVF conference on computer vision and pattern recognition}, 2024{\natexlab{b}}.

\bibitem[Lin(2004)]{lin2004rouge}
Chin-Yew Lin.
\newblock Rouge: A package for automatic evaluation of summaries.
\newblock In \emph{Text summarization branches out}, pages 74--81, 2004.

\bibitem[Liu et~al.(2023)Liu, Li, Wu, and Lee]{liu2023llava}
Haotian Liu, Chunyuan Li, Qingyang Wu, and Yong~Jae Lee.
\newblock Visual instruction tuning.
\newblock In \emph{NeurIPS}, 2023.

\bibitem[Liu et~al.(2024)Liu, Li, Wu, and Lee]{liu2024visual}
Haotian Liu, Chunyuan Li, Qingyang Wu, and Yong~Jae Lee.
\newblock Visual instruction tuning.
\newblock \emph{Advances in neural information processing systems}, 36, 2024.

\bibitem[Lu et~al.(2024)Lu, Liu, Zhang, Wang, Dong, Liu, Sun, Ren, Li, Yang, Sun, Deng, Xu, Xie, and Ruan]{lu2024deepseekvl}
Haoyu Lu, Wen Liu, Bo Zhang, Bingxuan Wang, Kai Dong, Bo Liu, Jingxiang Sun, Tongzheng Ren, Zhuoshu Li, Hao Yang, Yaofeng Sun, Chengqi Deng, Hanwei Xu, Zhenda Xie, and Chong Ruan.
\newblock Deepseek-vl: Towards real-world vision-language understanding, 2024.

\bibitem[Malla et~al.(2023)Malla, Choi, Dwivedi, Choi, and Li]{malla2023drama}
Srikanth Malla, Chiho Choi, Isht Dwivedi, Joon~Hee Choi, and Jiachen Li.
\newblock Drama: Joint risk localization and captioning in driving.
\newblock In \emph{Proceedings of the IEEE/CVF winter conference on applications of computer vision}, pages 1043--1052, 2023.

\bibitem[Mao et~al.(2023{\natexlab{a}})Mao, Ye, Qian, Pavone, and Wang]{agentdriver}
Jiageng Mao, Junjie Ye, Yuxi Qian, Marco Pavone, and Yue Wang.
\newblock A language agent for autonomous driving.
\newblock 2023{\natexlab{a}}.

\bibitem[Mao et~al.(2023{\natexlab{b}})Mao, Ye, Qian, Pavone, and Wang]{mao2023language}
Jiageng Mao, Junjie Ye, Yuxi Qian, Marco Pavone, and Yue Wang.
\newblock A language agent for autonomous driving.
\newblock \emph{arXiv preprint arXiv:2311.10813}, 2023{\natexlab{b}}.

\bibitem[Park et~al.(2024)Park, Lee, Kang, Choi, Park, Cho, Lee, and Kim]{park2024vlaad}
SungYeon Park, MinJae Lee, JiHyuk Kang, Hahyeon Choi, Yoonah Park, Juhwan Cho, Adam Lee, and DongKyu Kim.
\newblock Vlaad: Vision and language assistant for autonomous driving.
\newblock In \emph{Proceedings of the IEEE/CVF Winter Conference on Applications of Computer Vision}, pages 980--987, 2024.

\bibitem[Qian et~al.(2024)Qian, Chen, Zhuo, Jiao, and Jiang]{qian2024nuscenes}
Tianwen Qian, Jingjing Chen, Linhai Zhuo, Yang Jiao, and Yu-Gang Jiang.
\newblock Nuscenes-qa: A multi-modal visual question answering benchmark for autonomous driving scenario.
\newblock In \emph{Proceedings of the AAAI Conference on Artificial Intelligence}, pages 4542--4550, 2024.

\bibitem[Radford et~al.(2021)Radford, Kim, Hallacy, Ramesh, Goh, Agarwal, Sastry, Askell, Mishkin, Clark, et~al.]{radford2021learning}
Alec Radford, Jong~Wook Kim, Chris Hallacy, Aditya Ramesh, Gabriel Goh, Sandhini Agarwal, Girish Sastry, Amanda Askell, Pamela Mishkin, Jack Clark, et~al.
\newblock Learning transferable visual models from natural language supervision.
\newblock In \emph{International conference on machine learning}, pages 8748--8763. PMLR, 2021.

\bibitem[Reimers and Gurevych(2019)]{reimers-gurevych-2019-sentence}
Nils Reimers and Iryna Gurevych.
\newblock Sentence-{BERT}: Sentence embeddings using {S}iamese {BERT}-networks.
\newblock In \emph{Proceedings of the 2019 Conference on Empirical Methods in Natural Language Processing and the 9th International Joint Conference on Natural Language Processing (EMNLP-IJCNLP)}, pages 3982--3992, Hong Kong, China, 2019. Association for Computational Linguistics.

\bibitem[Shao et~al.(2024)Shao, Hu, Wang, Song, Waslander, Liu, and Li]{shao2024lmdrive}
Hao Shao, Yuxuan Hu, Letian Wang, Guanglu Song, Steven~L Waslander, Yu Liu, and Hongsheng Li.
\newblock Lmdrive: Closed-loop end-to-end driving with large language models.
\newblock In \emph{Proceedings of the IEEE/CVF Conference on Computer Vision and Pattern Recognition}, pages 15120--15130, 2024.

\bibitem[Sima et~al.(2023)Sima, Renz, Chitta, Chen, Zhang, Xie, Luo, Geiger, and Li]{sima2023drivelm}
Chonghao Sima, Katrin Renz, Kashyap Chitta, Li Chen, Hanxue Zhang, Chengen Xie, Ping Luo, Andreas Geiger, and Hongyang Li.
\newblock Drivelm: Driving with graph visual question answering.
\newblock \emph{arXiv preprint arXiv:2312.14150}, 2023.

\bibitem[Sun et~al.(2023)Sun, Fang, Wu, Wang, and Cao]{sun2023eva}
Quan Sun, Yuxin Fang, Ledell Wu, Xinlong Wang, and Yue Cao.
\newblock Eva-clip: Improved training techniques for clip at scale.
\newblock \emph{arXiv preprint arXiv:2303.15389}, 2023.

\bibitem[Team et~al.(2023)Team, Anil, Borgeaud, Wu, Alayrac, Yu, Soricut, Schalkwyk, Dai, Hauth, et~al.]{team2023gemini}
Gemini Team, Rohan Anil, Sebastian Borgeaud, Yonghui Wu, Jean-Baptiste Alayrac, Jiahui Yu, Radu Soricut, Johan Schalkwyk, Andrew~M Dai, Anja Hauth, et~al.
\newblock Gemini: a family of highly capable multimodal models.
\newblock \emph{arXiv preprint arXiv:2312.11805}, 2023.

\bibitem[Tian et~al.(2024{\natexlab{a}})Tian, Gu, Li, Liu, Hu, Wang, Zhan, Jia, Lang, and Zhao]{tian2024drivevlm}
Xiaoyu Tian, Junru Gu, Bailin Li, Yicheng Liu, Chenxu Hu, Yang Wang, Kun Zhan, Peng Jia, Xianpeng Lang, and Hang Zhao.
\newblock Drivevlm: The convergence of autonomous driving and large vision-language models.
\newblock \emph{arXiv preprint arXiv:2402.12289}, 2024{\natexlab{a}}.

\bibitem[Tian et~al.(2024{\natexlab{b}})Tian, Gu, Li, Liu, Zhao, Wang, Zhan, Jia, Lang, and Zhao]{DriveVLM}
Xiaoyu Tian, Junru Gu, Bailin Li, Yicheng Liu, Zhiyong Zhao, Yang Wang, Kun Zhan, Peng Jia, Xianpeng Lang, and Hang Zhao.
\newblock Drivevlm: The convergence of autonomous driving and large vision-language models.
\newblock \emph{arXiv preprint arXiv:2402.12289}, 2024{\natexlab{b}}.

\bibitem[Touvron et~al.(2023)Touvron, Lavril, Izacard, Martinet, Lachaux, Lacroix, Rozi{\`e}re, Goyal, Hambro, Azhar, et~al.]{touvron2023llama}
Hugo Touvron, Thibaut Lavril, Gautier Izacard, Xavier Martinet, Marie-Anne Lachaux, Timoth{\'e}e Lacroix, Baptiste Rozi{\`e}re, Naman Goyal, Eric Hambro, Faisal Azhar, et~al.
\newblock Llama: Open and efficient foundation language models.
\newblock \emph{arXiv preprint arXiv:2302.13971}, 2023.

\bibitem[Wang et~al.(2023)Wang, Lv, Yu, Hong, Qi, Wang, Ji, Yang, Zhao, Song, et~al.]{wang2023cogvlm}
Weihan Wang, Qingsong Lv, Wenmeng Yu, Wenyi Hong, Ji Qi, Yan Wang, Junhui Ji, Zhuoyi Yang, Lei Zhao, Xixuan Song, et~al.
\newblock Cogvlm: Visual expert for pretrained language models.
\newblock \emph{arXiv preprint arXiv:2311.03079}, 2023.

\bibitem[Xu et~al.(2020)Xu, Yang, Gong, Lin, Wu, Li, and Vasconcelos]{xu2020explainable}
Yiran Xu, Xiaoyin Yang, Lihang Gong, Hsuan-Chu Lin, Tz-Ying Wu, Yunsheng Li, and Nuno Vasconcelos.
\newblock Explainable object-induced action decision for autonomous vehicles.
\newblock In \emph{Proceedings of the IEEE/CVF Conference on Computer Vision and Pattern Recognition}, pages 9523--9532, 2020.

\bibitem[Xu et~al.(2023)Xu, Zhang, Xie, Zhao, Guo, Wong, Li, and Zhao]{xu2023drivegpt4}
Zhenhua Xu, Yujia Zhang, Enze Xie, Zhen Zhao, Yong Guo, Kenneth~KY Wong, Zhenguo Li, and Hengshuang Zhao.
\newblock Drivegpt4: Interpretable end-to-end autonomous driving via large language model.
\newblock \emph{arXiv preprint arXiv:2310.01412}, 2023.

\bibitem[Ye et~al.(2024)Ye, Xu, Ye, Yan, Hu, Liu, Qian, Zhang, and Huang]{ye2024mplug}
Qinghao Ye, Haiyang Xu, Jiabo Ye, Ming Yan, Anwen Hu, Haowei Liu, Qi Qian, Ji Zhang, and Fei Huang.
\newblock mplug-owl2: Revolutionizing multi-modal large language model with modality collaboration.
\newblock In \emph{Proceedings of the IEEE/CVF Conference on Computer Vision and Pattern Recognition}, pages 13040--13051, 2024.

\bibitem[Young et~al.(2024)Young, Chen, Li, Huang, Zhang, Zhang, Li, Zhu, Chen, Chang, et~al.]{young2024yi}
Alex Young, Bei Chen, Chao Li, Chengen Huang, Ge Zhang, Guanwei Zhang, Heng Li, Jiangcheng Zhu, Jianqun Chen, Jing Chang, et~al.
\newblock Yi: Open foundation models by 01. ai.
\newblock \emph{arXiv preprint arXiv:2403.04652}, 2024.

\bibitem[Yu et~al.(2024)Yu, Zhang, Yao, Dang, Chen, Lu, Cui, He, Liu, Chua, and Sun]{yu2024rlaifv}
Tianyu Yu, Haoye Zhang, Yuan Yao, Yunkai Dang, Da Chen, Xiaoman Lu, Ganqu Cui, Taiwen He, Zhiyuan Liu, Tat-Seng Chua, and Maosong Sun.
\newblock Rlaif-v: Aligning mllms through open-source ai feedback for super gpt-4v trustworthiness.
\newblock \emph{arXiv preprint arXiv:2405.17220}, 2024.

\bibitem[Zhai et~al.(2023)Zhai, Mustafa, Kolesnikov, and Beyer]{zhai2023sigmoid}
Xiaohua Zhai, Basil Mustafa, Alexander Kolesnikov, and Lucas Beyer.
\newblock Sigmoid loss for language image pre-training.
\newblock In \emph{Proceedings of the IEEE/CVF International Conference on Computer Vision}, pages 11975--11986, 2023.

\bibitem[Zhang et~al.(2023)Zhang, Dong, Wang, Cao, Xu, Ouyang, Zhao, Ding, Zhang, Duan, Zhang, Yan, Zhang, Li, Li, Chen, He, Zhang, Qiao, Lin, and Wang]{internlmxcomposer}
Pan Zhang, Xiaoyi Dong, Bin Wang, Yuhang Cao, Chao Xu, Linke Ouyang, Zhiyuan Zhao, Shuangrui Ding, Songyang Zhang, Haodong Duan, Wenwei Zhang, Hang Yan, Xinyue Zhang, Wei Li, Jingwen Li, Kai Chen, Conghui He, Xingcheng Zhang, Yu Qiao, Dahua Lin, and Jiaqi Wang.
\newblock Internlm-xcomposer: A vision-language large model for advanced text-image comprehension and composition.
\newblock \emph{arXiv preprint arXiv:2309.15112}, 2023.

\bibitem[Zheng et~al.(2024)Zheng, Chiang, Sheng, Zhuang, Wu, Zhuang, Lin, Li, Li, Xing, et~al.]{zheng2024judging}
Lianmin Zheng, Wei-Lin Chiang, Ying Sheng, Siyuan Zhuang, Zhanghao Wu, Yonghao Zhuang, Zi Lin, Zhuohan Li, Dacheng Li, Eric Xing, et~al.
\newblock Judging llm-as-a-judge with mt-bench and chatbot arena.
\newblock \emph{Advances in Neural Information Processing Systems}, 36, 2024.

\end{thebibliography}
}

\clearpage
\setcounter{page}{1}
\maketitlesupplementary
\appendix

% \clearpage
% \setcounter{page}{1}
% \noindent \begin{center} \textbf{\LARGE Appendix} \end{center}
% \appendix

The supplementary first provides more details of our dataset, covering data collection, augmentation, and key statistics, with sample presentations in Sec.~\ref{sec:more_datails_about_datasets}. We then outline the experimental setup, including datails of selected LVLMs, dataset division, assessment methods, and metric calculations in Sec.~\ref{sec:more_details_about_experiment_setup}. Next, we analyze the benchmark results in Sec.~\ref{sec:further_analysis_of_LVLM_results}, focusing on LVLMs' performance across different data domains and knowledge coverage. Finally, in Sec.~\ref{sec:Benefits_of_Driving_Knowledge}, we show that models fine-tuned with our dataset, in addition to nuScenes~\cite{caesar2020nuscenes}, produce safer, human-compliant trajectory planning, highlighting the impact of enriched driving knowledge.

\section{More Details about datasets}
\label{sec:more_datails_about_datasets}
In this section, we present detailed information on the data collection process (Sec. \ref{subsec: Data_Construction}), provide comprehensive statistics on the dataset (Sec. \ref{subsec: more_data_statistic}), and include additional data visualizations (Sec. \ref{subsec: more_data_example}).

\subsection{Data Construction}
\label{subsec: Data_Construction}

\subsubsection{Driving Handbook Data Collection}
Driving handbooks are highly structured and comprehensive resources, covering laws, regulations, techniques, safety, and more. To extract unordered data from these documents, we first utilize layout detection and Optical Character Recognition (OCR) technology to extract data blocks and the text within these blocks. Specially, we use LayoutLMv3~\cite{huang2022layoutlmv3} as our layout detection and $PaddleOCR$ as our OCR toolkit. Next, we utilize DBSCAN~\cite{DBSCAN} to cluster blocks to let each column in a multi-column PDF falls into the same cluster. After that, we sequence the data blocks based on the position and which column it's in. Finally we filter out duplicate and irrelevant data.

\subsubsection{Driving Road Data Collection}
In the CARLA, the position and rotation of an actor are represented by carla.Location and carla.Rotation, respectively. The carla.Location format is given as (x, y, z) in a rectangular coordinate system, measured in meters. While CARLA is based on the UE4, the carla.Rotation differs in its definition, using (pitch, yaw, roll) to describe orientation. Specifically, pitch represents rotation around the Y-axis, yaw represents rotation around the Z-axis, and roll represents rotation around the X-axis, all measured in degrees.

During the Camera Data Collection stage, we captured the position and rotation of the ego vehicle and traffic signs. Subsequently, in the Data Annotation stage, images containing traffic signs were identified based on the position, distance, and rotation of the traffic signs relative to the ego vehicle. The distance is calculated as follows:
\begin{equation}
\text{distance} = \sqrt{(x_{sign}-x_{ego})^{2} + (y_{sign}-y_{ego})^{2}} 
\end{equation}
where, $x_{sign}$, $y_{sign}$ represent the location of the traffic sign and $x_{ego}$, $y_{ego}$ represent the location of the ego vehicle.

In CARLA, the yaw ranges from \(-180^\circ\) to \(180^\circ\). A yaw of \(0^\circ\) indicates that the actor is facing the positive X-axis. A positive yaw value corresponds to a counterclockwise rotation (left turn), while a negative yaw value indicates a clockwise rotation (right turn).

The locational angle of the traffic sign relative to the ego vehicle is calculated as:
\begin{equation}
\text{yaw}_{loc} = \left(\arctan \left(\frac{y_{sign} - y_{ego}}{x_{sign} - x_{ego}}\right) \times \frac{180}{\pi}\right) - \text{yaw}_{ego}
\end{equation}
where, $\text{yaw}_{ego}$ represent the rotation around the Z-axis of the ego vehicle.

Then the $\text{yaw}_{loc}$ is normalized to the range \([-180^\circ, 180^\circ]\) by:
\begin{equation}
\text{yaw}_{loc} = \text{yaw}_{loc} \bmod 360
\end{equation}

\begin{equation}
\text{yaw}_{loc} = \begin{cases} 
\text{yaw}_{loc} - 360 & \text{if } \text{yaw}_{loc} > 180 \\
\text{yaw}_{loc} & \text{otherwise}
\end{cases}
\end{equation}

The directional angle of the traffic sign relative to the ego vehicle is calculated as follows:
\begin{equation}
\text{yaw}_{rot} = \text{yaw}_{sign} - \text{yaw}_{ego}
\end{equation}
where, $\text{yaw}_{sign}$ represent the rotation around the Z-axis of the traffic sign.

Then the $\text{yaw}_{rot}$ is normalized to the range \([-180^\circ, 180^\circ]\) by the same way as above.

After testing, we finally determined that the $\text{yaw}_{loc}$, $\text{distance}$ and $\text{yaw}_{rot}$ are respectively $([-35^\circ, 35^\circ])$, $((0, 30m])$, $([-180^\circ, -160^\circ])\bigcup([160^\circ, 180^\circ])$, and judged in turns.

\subsubsection{Data Augmentation.}
To enhance the diversity of data and expand the scale of the dataset, we employed GPT-4o model to augment data. To ensure the enhanced data is well-structured, we divide the item into the question stem, options, and explanation sections and enhance them separately. After that enhancaed sections are combined to form the enhanced question-answer pair. Prompt we used are shown in Tab. \ref{tab:prompt}.

\begin{table}[t]
\centering
\caption{\textbf{Key statistics of IDKB.} 
The percentage represents the proportion of the data relative to the total questions.}
\resizebox{\linewidth}{!}{
\begin{tabular}{lc}
\toprule Statistics & Number \\
\midrule Total Questions & 1016959 \\
    \hspace{5mm}\** Driving Handbook Data &  50501 (5.0\%) \\
    \hspace{5mm}\** Driving Test Data &  854067 (84.0\%) \\
    \hspace{5mm}\** Driving Road Data &  112388 (11.1\%) \\
Train:Test & 931553:85406 \\
\midrule Multi-Choice Questions (MCQ) & 830057 (81.5\%) \\
    \hspace{5mm}\** Single answer &  754595 (74.1\%) \\
    \hspace{5mm}\** Multiple answers &  74926 (7.4\%) \\
Open QA Questions (QA) & 187435 (18.5\%) \\
\midrule Questions with an explanation & 66645 (6.5\%) \\
\midrule Total images & 493699 \\
    \hspace{5mm}\** Images in the question & 491710 \\
    \hspace{5mm}\** Images in the options & 1989 \\
Questions with multiple images & 4560 (0.4\%)\\
\midrule Average question length & 26.77 \\
Average option length & 11.18 \\
\bottomrule
\end{tabular}}
\label{tab:key_statistics}
\end{table}

\begin{table}[t]
\centering
\caption{\textbf{Country distribution of IDKB.}}
\resizebox{\linewidth}{!}{
\begin{tabular}{lcc}
\toprule Country & Annotation Language & Number \\
\midrule America & English & 226817 \\
China & Chinese (Simplified / Traditional) & 204273 \\
Italy & Italian & 183869 \\
German & Germany / English & 92254 \\
Spain & Spanish & 63553\\
Canada & English & 55301\\
UK & English & 53958 \\
France & French & 35718 \\
Korea & Korean / English & 24284 \\
Australia & English & 21585 \\
Singapore & English & 15210 \\
Japan & Japanese & 14162 \\
Ireland & English & 11264 \\
New Zealand & English & 9555\\
India & English & 5153\\
\bottomrule
\end{tabular}}
\label{tab:country_distribution}
\end{table}

\begin{table*}[t]
\centering
\caption{\textbf{Model architecture of 15 LVLMs evaluated on IDKB.}}
\begin{tabular}{lccc}
\toprule
Models                                                   & Parameters & Vision Encoder    & LLM                \\ \midrule
GPT-4o \cite{achiam2023gpt}  & -          & -                 & -                  \\
Gemini-1.5-flash \cite{team2023gemini} & -          & -                 & -                  \\
VisualGLM-6B \cite{du2022glm}         & 8B         & EVA-CLIP          & ChatGLM-6B         \\
Qwen-VL-chat \cite{Qwen-VL}           & 9.6B       & CLIP ViT-G/16     & QWen-7B            \\
CogVLM \cite{wang2023cogvlm}          & 17B        & EVA-CLIP-E        & Vicuna-v1.5-7B     \\
BLIP2 \cite{li2023blip2bootstrappinglanguageimagepretraining} & 12.1B & EVA-CLIP ViT-G/14 & Flan-T5-XXL \\
LLaVA-v1.5-7B \cite{liu2023llava}      & 7.2B       & CLIP ViT-L/14     & Vicuna-v1.5-7B     \\
XComposer \cite{internlmxcomposer}    & 8B         & EVA-CLIP-G        & InternLM-7B        \\
ShareGPT4V-7B \cite{chen2023sharegpt4v} & 7.2B       & CLIP ViT-L/14     & Vicuna-v1.5-7B     \\
mPLUG-Owl2 \cite{ye2024mplug}          & 8.2B       & CLIP ViT-L/14     & LLaMA2-7B          \\
MiniCPM-Llama3-V2.5 \cite{yu2024rlaifv}& 8B         & SigLip            & Llama3-8B-Instruct \\
XComposer2 \cite{internlmxcomposer2} & 7B         & CLIP ViT-L/14     & InternLM2-7B       \\
DeepSeek-VL-7B \cite{lu2024deepseekvl}  & 7.3B       & SAM-B \& SigLIP-L & DeekSeek-7B        \\
Yi-VL-6B \cite{ai2024yi}             & 6.6B       & CLIP ViT-H/14     & Yi-6B              \\
Monkey-chat \cite{li2023monkey}    & 9.8B       & CLIP-ViT-BigHuge  & Qwen-7B           \\ \bottomrule
\end{tabular}
\label{tab:LVlMs_details}
\end{table*}

\subsection{Data Statistics}
\label{subsec: more_data_statistic}
Tab.~\ref{tab:key_statistics} provides a summary of detailed statistics for IDKB, including numbers on dataset split, question types, image position, average text length, and other relevant statistics. Note that we calculate average question/option length on the unaugmented data, and special tokenizers are utilized to count words for Chinese, Korean and Japanese text. Tab.~\ref{tab:country_distribution} presents detailed numbers on country distribution. The annotation language of each country's data is also listed. 

To obtain a statistical view of the knowledge domains within IDKB, we designed specific prompts and employed proprietary LVLMs to classify all the questions into four major categories based on their semantics. Here we provide detailed definitions of the four kinds of knowledge domains: 
\begin{itemize}
    \item Traffic Laws and Regulations:
This category focuses exclusively on the statutory and legal aspects of driving. It includes questions about state and local traffic laws, understanding of legal driving ages, alcohol and drug influence laws, and penalties for traffic violations. This category ensures that drivers are aware of legal obligations and the consequences of their actions on the road, which can be a substantial part of a driving test.
\item Road Signs, Signals, and Lane Markings:
This category is dedicated to testing knowledge of road signs (stop, yield, pedestrian crossing, etc.), traffic signals, and different lane markings. This includes interpreting what various signs and markings mean and how to respond to them while driving. The category serves to evaluate a driver’s ability to navigate roads safely and appropriately, understanding that correct responses to signs and signals are crucial for safe driving.
\item Vehicle Control and Driving Techniques:
Here, the focus is on practical driving skills and vehicle management. Questions could include scenarios dealing with vehicle control in adverse weather conditions, the correct use of mirrors, turning and passing maneuvers, and parking techniques. This category tests a driver’s hands-on ability to operate a vehicle safely and maintain control in everyday driving situations as well as in unexpected conditions.
\item Driver Responsibility and Defensive Driving:
This category assesses knowledge related to driver’s behavior and responsibilities. It includes defensive driving techniques, the importance of seat belts, methods to prevent distracted driving, and how to act responsibly in various driving situations (school zones, neighborhoods). Questions about accident procedures, emergency responses, and first aid may also be included. It underscores the importance of personal responsibility and proactive safety measures in driving.
\end{itemize}

\subsection{More Data Examples}
\label{subsec: more_data_example}
\begin{figure*}[t]
  \includegraphics[width=\linewidth]{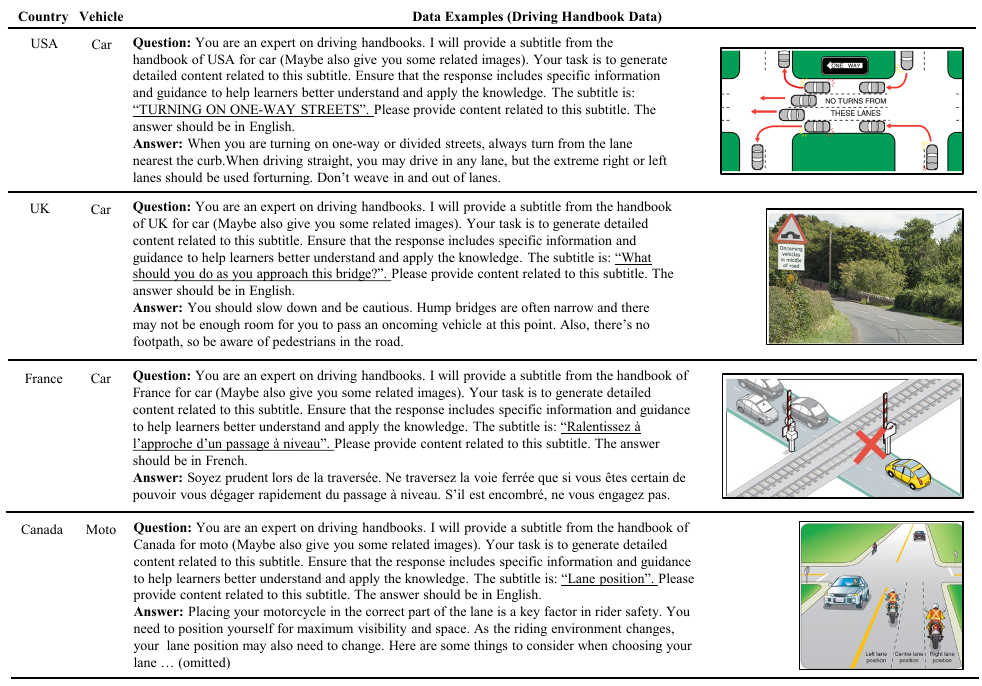}
  \caption{\textbf{More Examples of Driving Handbook Data.}}
\label{fig:handbook_data_example}
\end{figure*}

\begin{figure*}[t]
  \includegraphics[width=\linewidth]{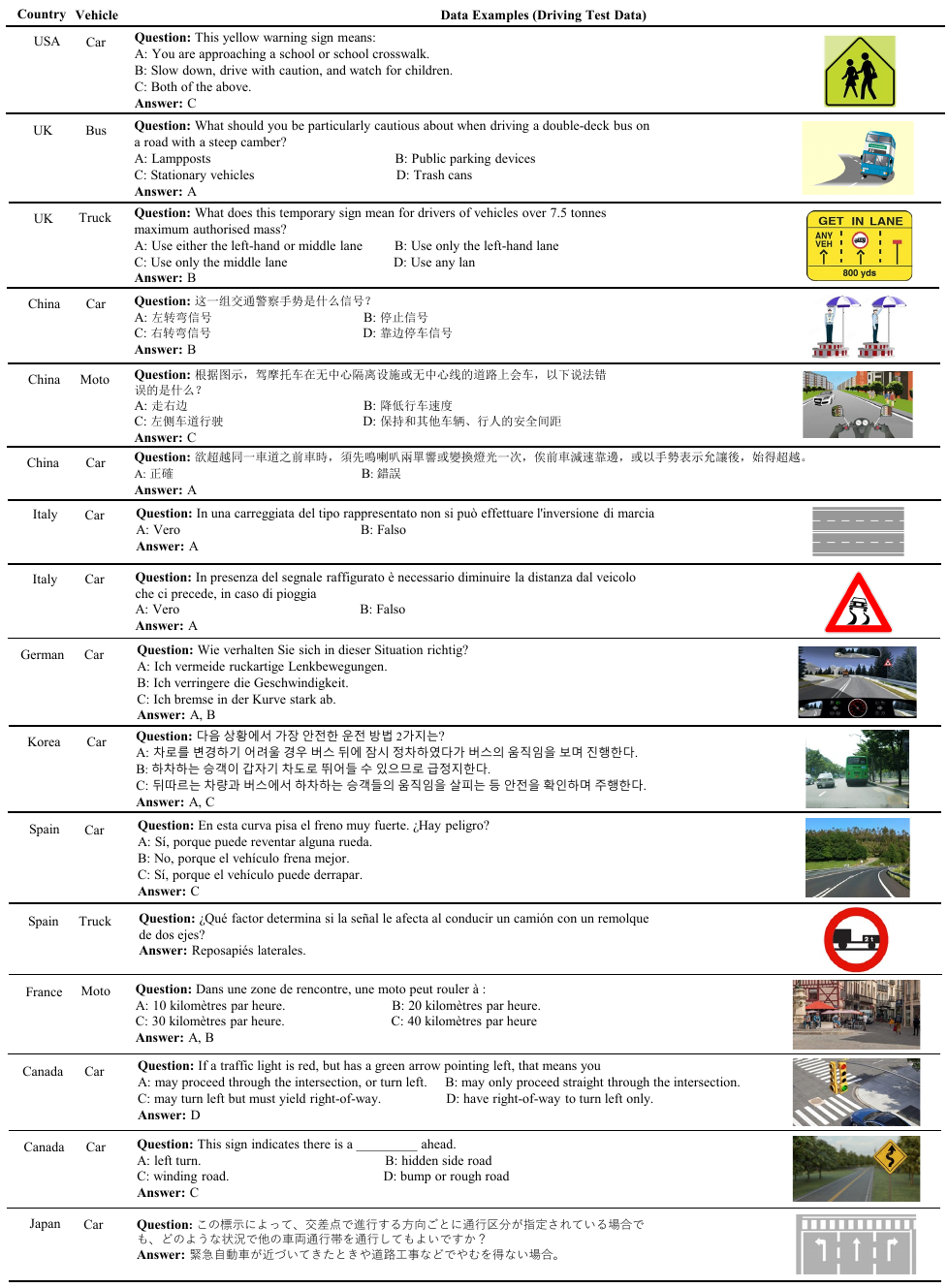}
  \caption{\textbf{More Examples of Driving Test Data.}}
\label{fig:test_data_example}
\end{figure*}

\begin{figure*}[t]
  \includegraphics[width=\linewidth]{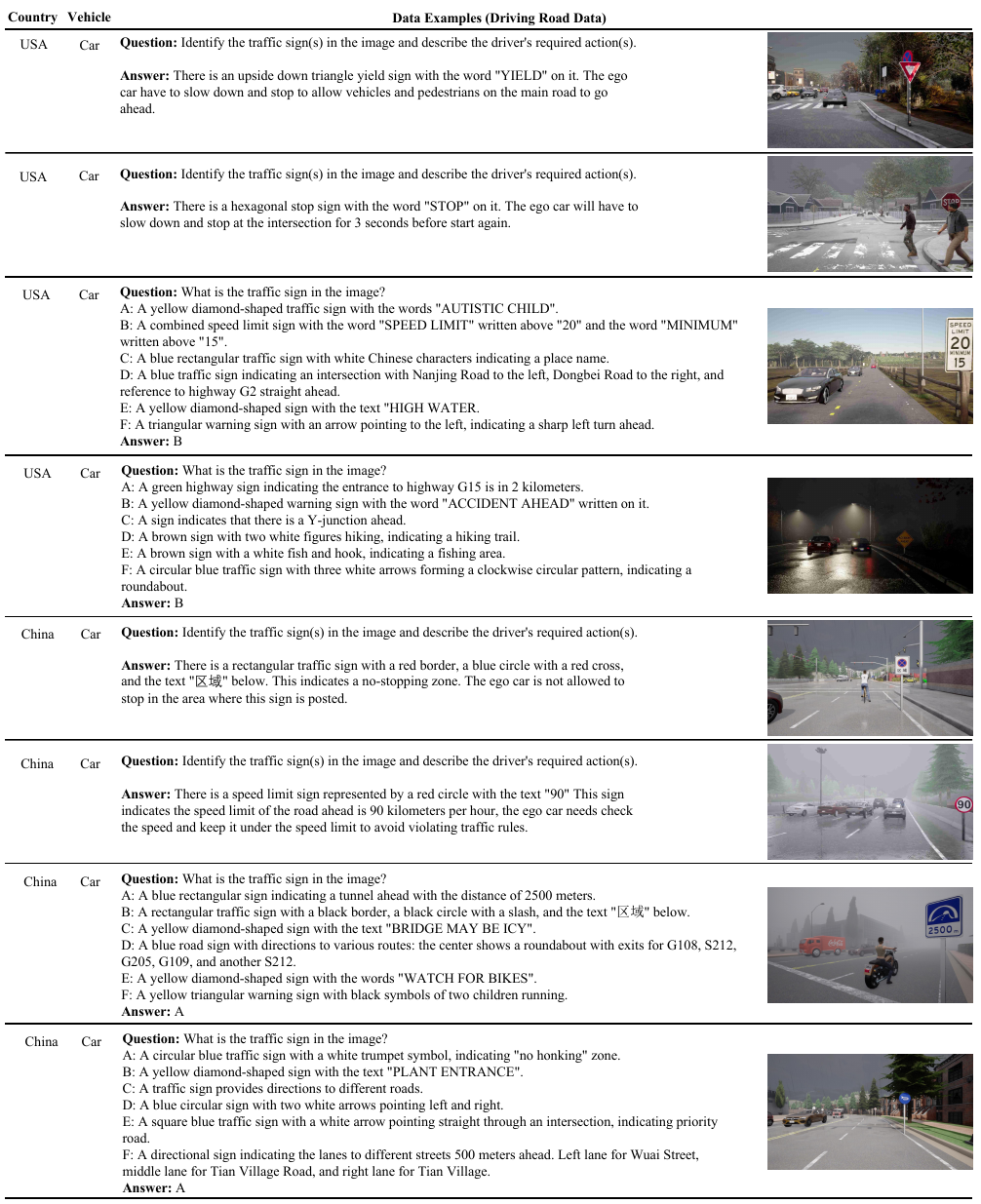}
  \caption{\textbf{More Examples of Driving Road Data.}}
\label{fig:road_data_example}
\end{figure*}

To provide a more intuitive overview of our IDKB dataset, we present additional data examples in this section, as illustrated in Fig.\ref{fig:handbook_data_example}, \ref{fig:test_data_example} and \ref{fig:road_data_example}, for Driving Handbook Data, Driving Test Data, and Driving Road Data respectively.

\begin{table*}[h]
\centering
\caption{Prompts used for data enhancement and model inference.}
% \resizebox{\linewidth}{!}{
\begin{tabular}{lp{10cm}}
\toprule Task & Prompt \\
\midrule Data Augmentation &  \\
    \hspace{5mm}\** Text Enhancement &  Translate and generate sentence or paragraph that maintain the same meaning. Make sure your sentences are smooth. If the sentence is too long, simplify them. Don't change the sentence type. If the sentence is incomplete, do not complete it and do not raise error. Provide 3 version each time, each version should be different. The beginning and end of the provided sentences or paragraph is marked with two dollar signs(\$\$). All the content between two dollar signs should be treated as sentence or paragraph that need to be translated and generated under the requirement mentioned before. Input format will be \$\$xxxxx.......\$\$. Between each version put four dollar sign(\$). Do not add version serial number. Only the translated and generated version needs to be output, not the original sentence or paragraph. Output language should be English.  Output format should be: xxxx \$\$\$\$ xxxx \$\$\$\$ xxxx Here is the sentence or paragraph: \{ $Original \ Text$ \} \\
    \hspace{5mm}\** Image to Text Enhancement &  Describe the image \\

\midrule Inference &  \\
    \hspace{5mm}\**  Single Answer MCQ Questions &  \\

    \hspace{10mm}\*\*** Driving Test Data & Think carefully and tell me which answer is correct.Format should be: Optionx (Option need to be output) \newline Answer this \{ $Country$ \}\ \ \{ $Vehicle \ Type$ \} driving test question. \newline Question: \{ $Question$ \} \\
    
    \hspace{10mm}\*\*** Driving Road Data & Think carefully and tell me which answer is correct.Format should be: Optionx (Option need to be output) \newline Assume you are driving a car in \{ $Country$ \}. \newline Question: \{ $Question$ \} \\

    \hspace{5mm}\** Multiple Answer MCQ Questions &  \\
    \hspace{10mm}\*\*** Driving Test Data & A multiple choice question will be provided. Think carefully and tell me the answer is correct. Use only comma between chosen options. Format should be: Optionx, Optionx, Optionx, ...(if needed and Option need to be output) \newline Answer this \{ $Country$ \}\ \ \{ $Vehicle \ Type$ \} driving test question. \newline Question: \{ $Question$ \} \\
    
    \hspace{10mm}\*\*** Driving Road Data & A multiple choice question will be provided. Think carefully and tell me the answer is correct. Use only comma between chosen options. Format should be: Optionx, Optionx, Optionx, ...(if needed and Option need to be output) \newline Assume you are driving a car in \{ $Country$ \}. \newline Question: \{ $Question$ \} \\
\bottomrule
\end{tabular}
% }
\label{tab:prompt}
\end{table*}

\section{More Details about Experiment Setup}
\label{sec:more_details_about_experiment_setup}
In this section, we provide additional information on the selected LVLMs in Sec.~\ref{subsec:LVLMs Model Details}, outline the evaluation methods for multiple-choice and question-and-answer questions in Sec.~\ref{subsec:Evaluation_Details}, and explain the computation of the three overall metrics in Sec.~\ref{subsec:overall_metrics}. Then in Sec.~\ref{subsec:prompt} we introduce the prompts we used in the paper.

\subsection{LVLMs Model Details}
\label{subsec:LVLMs Model Details}
Tab.~\ref{tab:LVlMs_details} provides details of the LVLMs used in this paper, including their parameter sizes, visual encoders, and LLMs. The selected open-source LVLMs have parameter sizes ranging from 6.6B to 12.1B, featuring 4 different vision encoders and 11 distinct LLMs. Finetuning and inference times also vary across models. For instance, finetuning Qwen-VL-chat with LoRA~\cite{hu2021lora} on our dataset, using 8 NVIDIA A40 GPUs with a maximum token length of 2048, took 66 hours and required approximately 24GB of memory.

\begin{algorithm}
\caption{Extract and Evaluate Answers}
\begin{algorithmic}[1]
\STATE \textbf{Input:} A string containing potential answers $O$, correct answer(s) $Ans$
\STATE \textbf{Output:} \text{True if the answer is correct, otherwise False}

\STATE \textbf{Step 1: Extract Answers}
\IF{The input contains a pattern like 'Option + letter'}
    \STATE Extract the letter(s) following 'Option' from $O$
\ELSIF{The input does not match 'Option + letter' but contains 'is + letter'}
    \STATE Extract the letter(s) following 'is' from $O$
\ELSE
    \STATE Extract individual capital letters from $O$
\ENDIF

\STATE \textbf{Step 2: Evaluate Correctness}
\IF{The length of $Ans$ is 1 (single answer)}
    \STATE Compare the extracted answer(s) with $Ans$
\ELSE
    \IF{The number of extracted answers does not match the number of options in $Ans$}
        \STATE \textbf{Return} False
    \ELSE
        \STATE Compare the extracted answers with $Ans$
    \ENDIF
\ENDIF
\STATE \textbf{Return} True if the answers match, otherwise False
\end{algorithmic}
\label{code:extract_and_evaluate_answers}
\end{algorithm}

\subsection{Evaluation Details}
\label{subsec:Evaluation_Details}
\noindent\textbf{Multiple-Choice Questions}
For multiple-choice questions, we use regular expressions to extract answers according to the logic outlined in Algorithm \ref{code:extract_and_evaluate_answers}. For instruction-following, we simply check whether the LVLM's output contains the pattern ``Option + letter''.

\noindent\textbf{Question-and-Answer}
For question-and-answer, we employ ROUGE~\cite{lin2004rouge} and SEMScore~\cite{aynetdinov2024semscoreautomatedevaluationinstructiontuned} to measure similarity between LVLM outputs and the reference answers. ROUGE evaluates N-gram overlap between outputs and the reference answers. For each question we consider enhanced answers as reference too to duel with free-form nature of LVLM outputs, especially for ROUGE. We compute ROUGE-1 and ROUGE-L between the output and each reference answer, and then select the maximum score for both ROUGE-1 and ROUGE-L.

For SEMScore, we first utilize sentence transformer~\cite{reimers-gurevych-2019-sentence} to embed LVLM outputs and the reference answers. Specifically we use ``paraphrase-multilingual-mpnet-base-v2'' as our model. Then we measured cosine distance between embeds which represents the difference between output and the reference answers in semantic level. Similar to Rouge, we take the maximum SEMScore as the SEMScore for the output.

\subsection{Overall Metrics}
\label{subsec:overall_metrics}
The overall score for MCQs is calculated as the weighted accuracy of both single-choice and multiple-choice questions, with the quantity of each question type used as the weighting factor. The value of instruction-following is excluded from this calculation and is used only as a reference.

For QAs, the overall score is a weighted combination of Rouge-1, Rouge-L, and SEMScore, with weights of 0.15, 0.15, and 0.70, respectively. Given the free-form nature of LVLM outputs, semantically similar responses may receive low ROUGE scores due to limited N-gram overlap. To address this, we assign a higher weight to SEMScore to better capture human-like evaluation.

The Test Data Score and Road Data Score are calculated as the averages of the overall MCQ and QA scores from the driving test data and driving road data, respectively. The IDKB score is then derived as the average of the Test Data Score and Road Data Score.

\subsection{Prompt}
\label{subsec:prompt}
We include the country, vehicle type, and question type in our prompt to provide background knowledge to the LVLMs. Additionally, The prompt also includes a detailed description of the required output format to ensure the responses meet the specified standards. Prompt we used to evaluate LVLMs are shown in Tab. \ref{tab:prompt}.

\section{Further Analysis of LVLM Results}
\label{sec:further_analysis_of_LVLM_results}
In this section, we analyze the entire test set, including both the Driving Test Data and Driving Road Data. We examine the performance of all evaluated models, focusing on overall performance (Sec.~\ref{subsec: eval_test_set}), performance across different countries (Sec.~\ref{subsec: eval_by_country}), languages (Sec.~\ref{subsec: eval_by_language}), and vehicle types (Sec.~\ref{subsec: eval_by_vehicle_type}), as well as performance across various knowledge domains (Sec.~\ref{subsec: eval_by_knowledge}).

\subsection{Evaluate LVLMs Performance on test set}
\label{subsec: eval_test_set}
\begin{figure*}[h]
  \includegraphics[width=\linewidth]{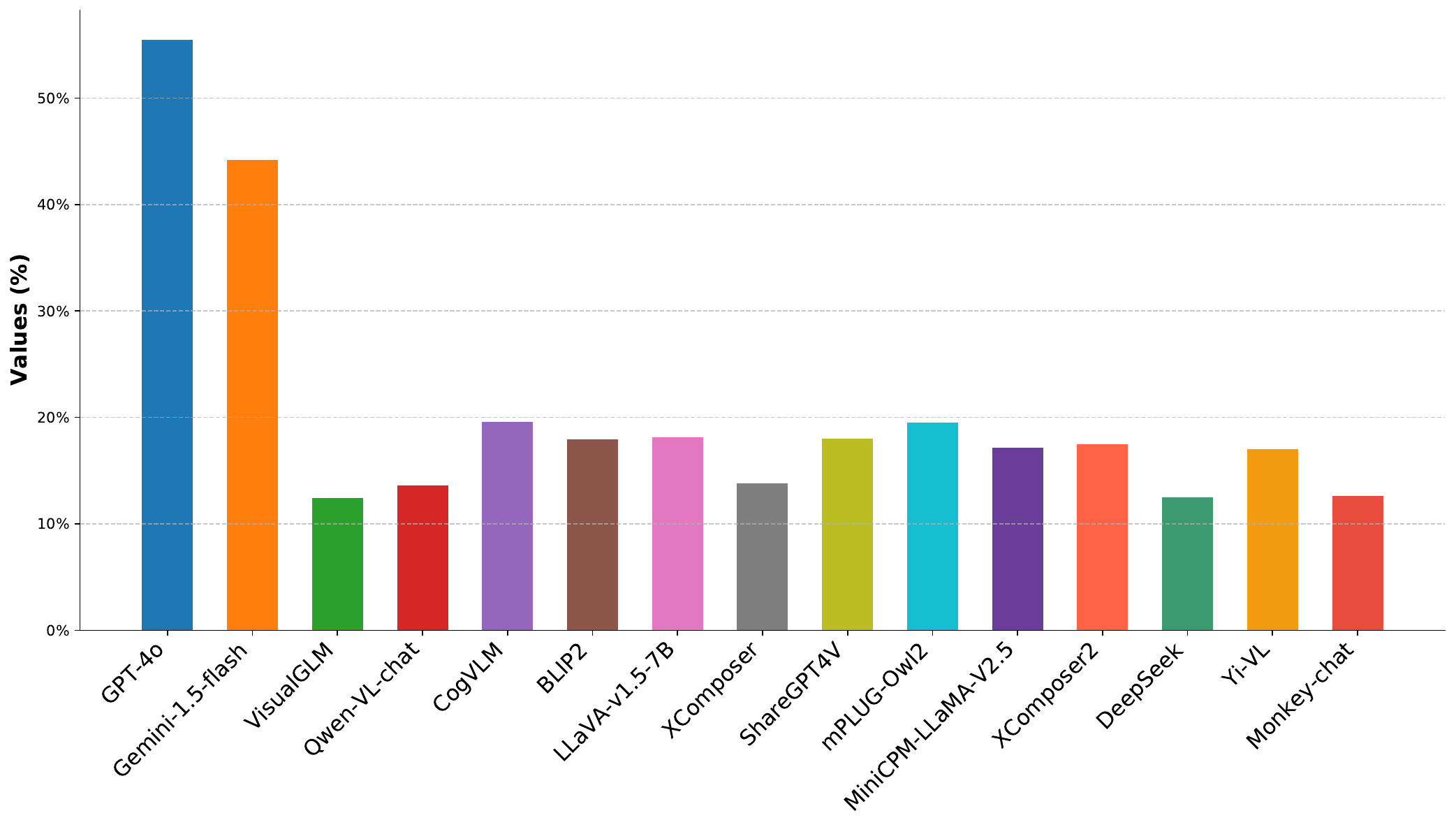}
  \caption{\textbf{Model Accuracy for Multiple-Choice Questions on test set.}}
\label{fig:more_analysis_mcq_overall_test}
\end{figure*}

\begin{figure*}[h]
  \includegraphics[width=\linewidth]{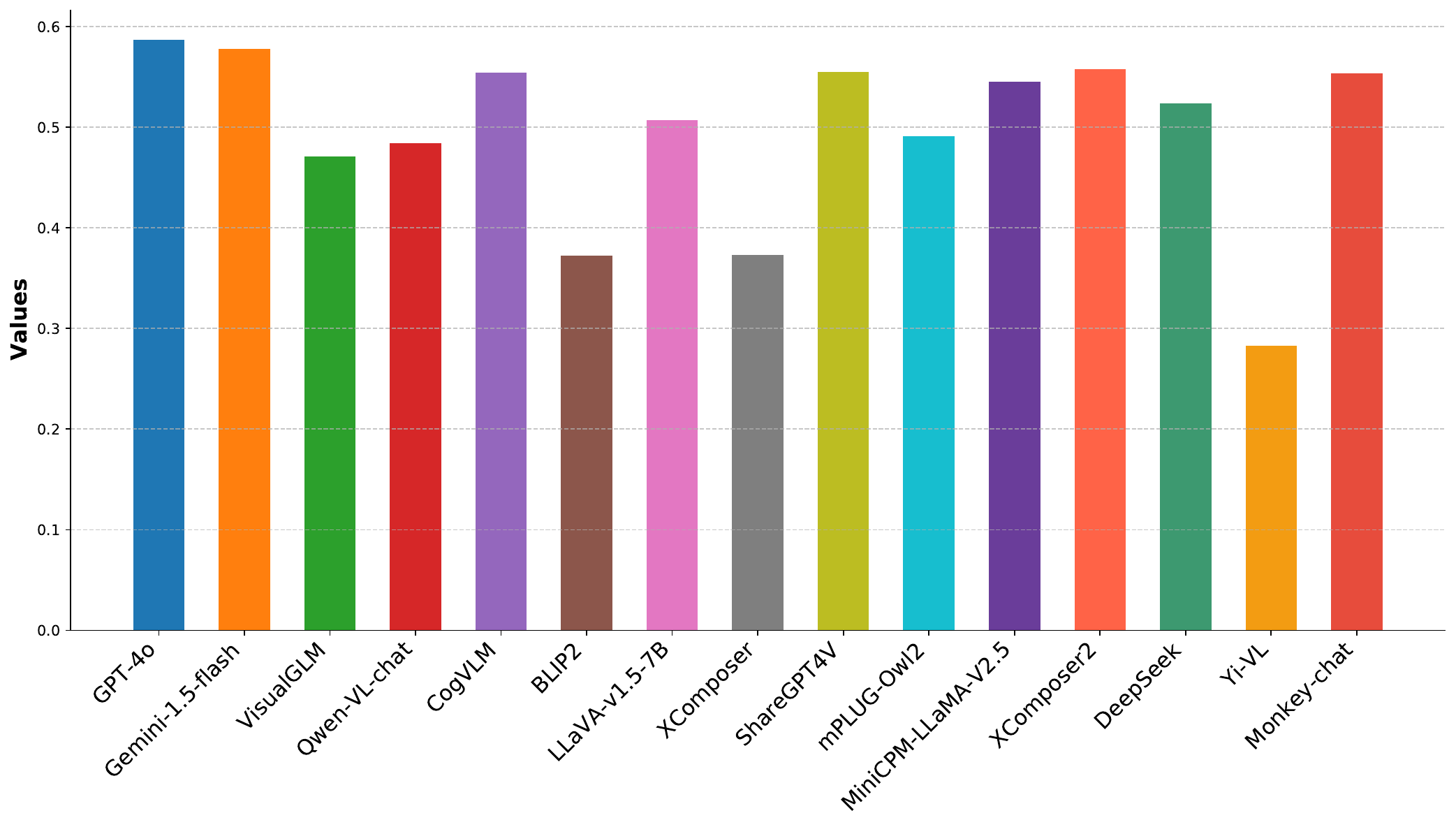}
  \caption{\textbf{Model Score for Question-and-Answer Questions on test set.}}
\label{fig:more_analysis_qa_overall_test}
\end{figure*}

As shown in Fig.~\ref{fig:more_analysis_mcq_overall_test} and Fig.~\ref{fig:more_analysis_qa_overall_test}, for MCQ questions, the proprietary model significantly outperformed the open-source models, with the accuracy gap between the best and worst open-source performers (CogVLM vs. VisualGLM) being nearly twofold. However, for QA questions, the gap is less pronounced, with many open-source models performing comparably to proprietary models.

\subsection{Evaluate LVLMs Performance by Country}
\label{subsec: eval_by_country}
We provide the performance of different models on questions from different countries in the Fig.~\ref{fig:more_analysis_mcq_country} and Fig.~\ref{fig:more_analysis_qa_country}. MCQ questions involve 14 countries, while QA questions cover 3 countries.
For QA questions, the performance gap between models is consistent across the three countries. However, for questions from India, all models scored lower compared to the other two countries.
In the case of MCQ questions, we observed significant variations in LVLM performance across countries. LLaVA-v1.5-7B, XComposer, and ShareGPT4V performed exceptionally well on Japanese questions, significantly outperforming other models. LLaVA-v1.5-7B and ShareGPT4V also delivered strong results on Italian questions. For UK questions, GPT-4 achieved an accuracy exceeding 80\%, which is particularly impressive. However, on questions from Japan, Korea, and Spain, some LVLMs had an accuracy of 0. This disparity highlights the varying capabilities of different LVLMs across countries and underscores the value of our dataset's diversity in providing a comprehensive evaluation

\subsection{Evaluate LVLMs Performance by Language}
\label{subsec: eval_by_language}
\begin{figure*}[h]
  \includegraphics[width=\linewidth]{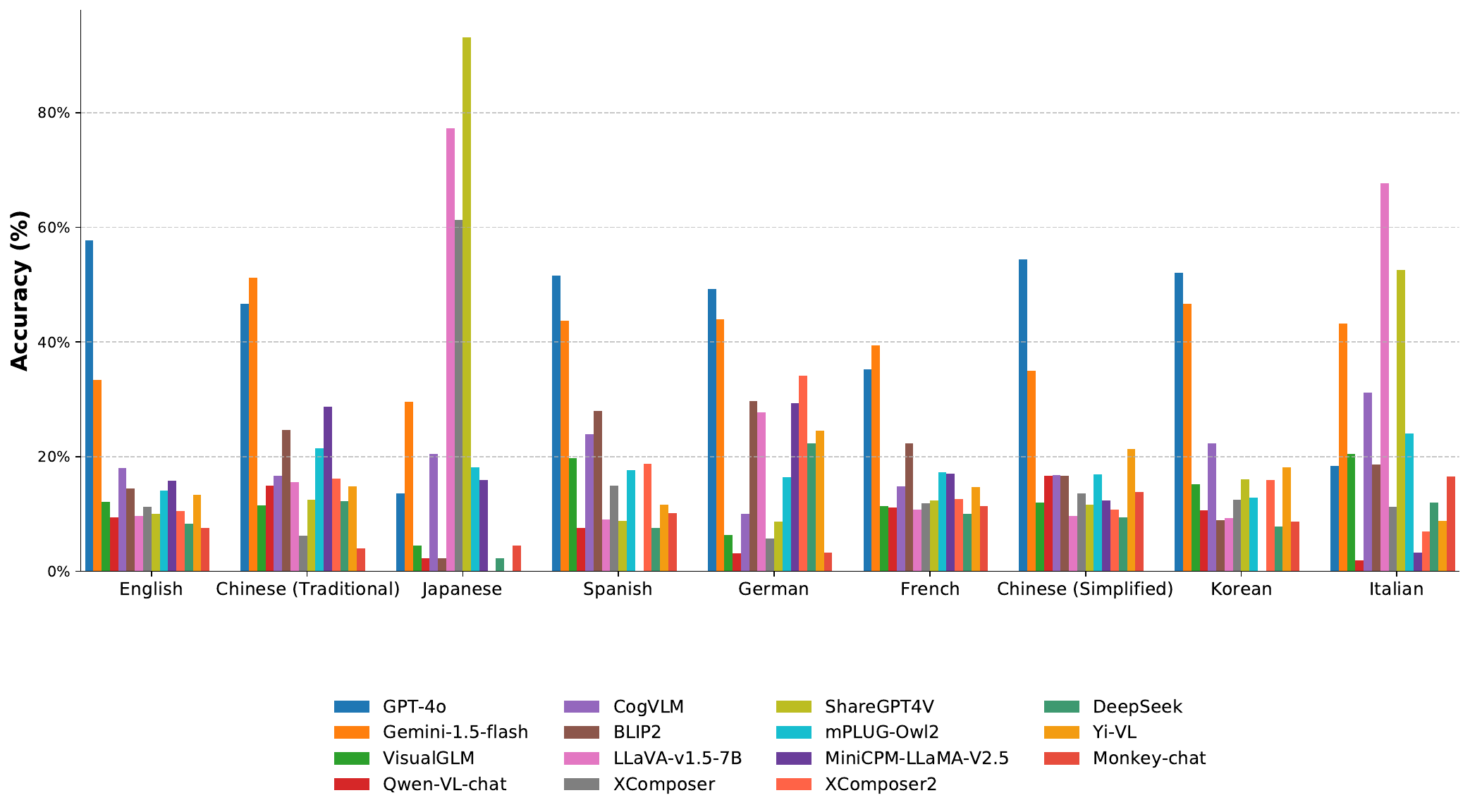}
  \caption{\textbf{Model Accuracy for Multiple-Choice Questions by Language.}}
\label{fig:more_analysis_mcq_language}
\end{figure*}

Fig.~\ref{fig:more_analysis_mcq_language} illustrate the performance of various models on questions across different languages. Since the QA questions in the test set are exclusively in English, this analysis focuses solely on the MCQ questions. The performance disparity between models is more pronounced for minority languages compared to English. LLaVA-v1.5-7B, XComposer, and ShareGPT4V performed exceptionally well on Japanese questions, whereas GPT-4 faced some challenges in this area. CogVLM and BLIP2 showed strong performance across several minority languages, including Spanish, German, French, Korean, and Italian. In contrast, MiniCPM-LLaMA-V2.5 struggled with minority languages, achieving a correct rate of 0 on Spanish and Korean questions, although it performed well on Traditional Chinese. These results demonstrate that models exhibit different strengths depending on the language and region. While some models excel in specific languages or regions, they may underperform in others. This variation underscores the importance of evaluating models across a diverse set of languages and regions to fully understand their capabilities and limitations.

\subsection{Evaluate LVLMs Performance by Vehicle Type}
\label{subsec: eval_by_vehicle_type}
\begin{figure*}[h]
  \includegraphics[width=\linewidth]{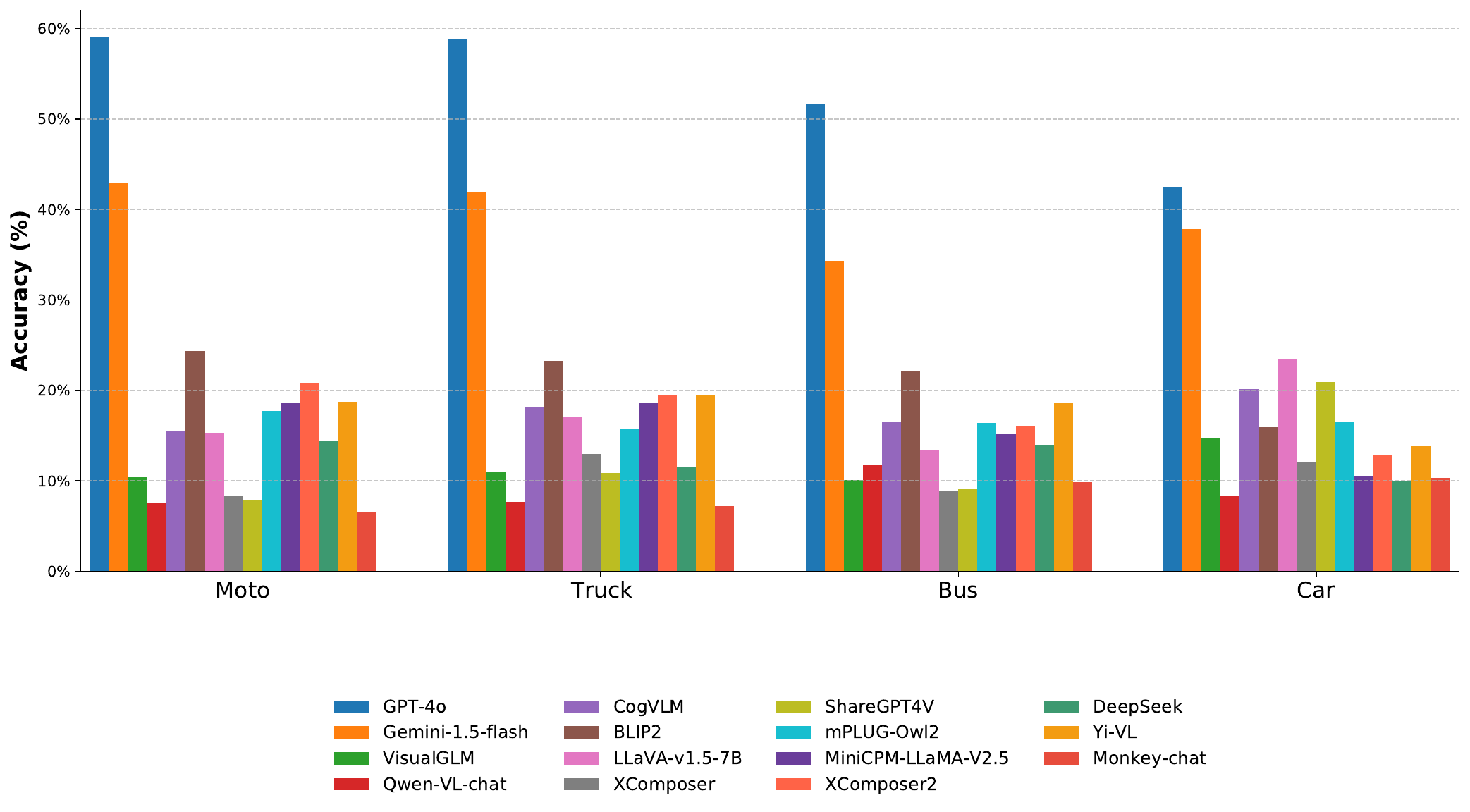}
  \caption{\textbf{Model Accuracy for Multiple-Choice Questions by Vehicle Type.}}
\label{fig:more_analysis_mcq_vehicle_type}
\end{figure*}

Since all QA questions in the test set are of the ``car'' type, our analysis concentrates on the MCQ questions. Fig.~\ref{fig:more_analysis_mcq_vehicle_type} illustrates the performance of various models across different vehicle types. Among the open-source models, BLIP2 demonstrates the strongest performance in the motor, truck, and bus categories, while LLaVA-v1.5-7B excels in the car category. Notably, Qwen-VL-chat shows consistent but average performance across motor, truck, and car categories. Interestingly, the models tend to perform better on questions related to motor, truck, and bus than on car-related questions, suggesting that the car category poses unique challenges that may require further exploration.

\subsection{Evaluate LVLMs Performance by Knowledge Coverage}
\label{subsec: eval_by_knowledge}
\begin{figure*}[h]
  \includegraphics[width=\linewidth]{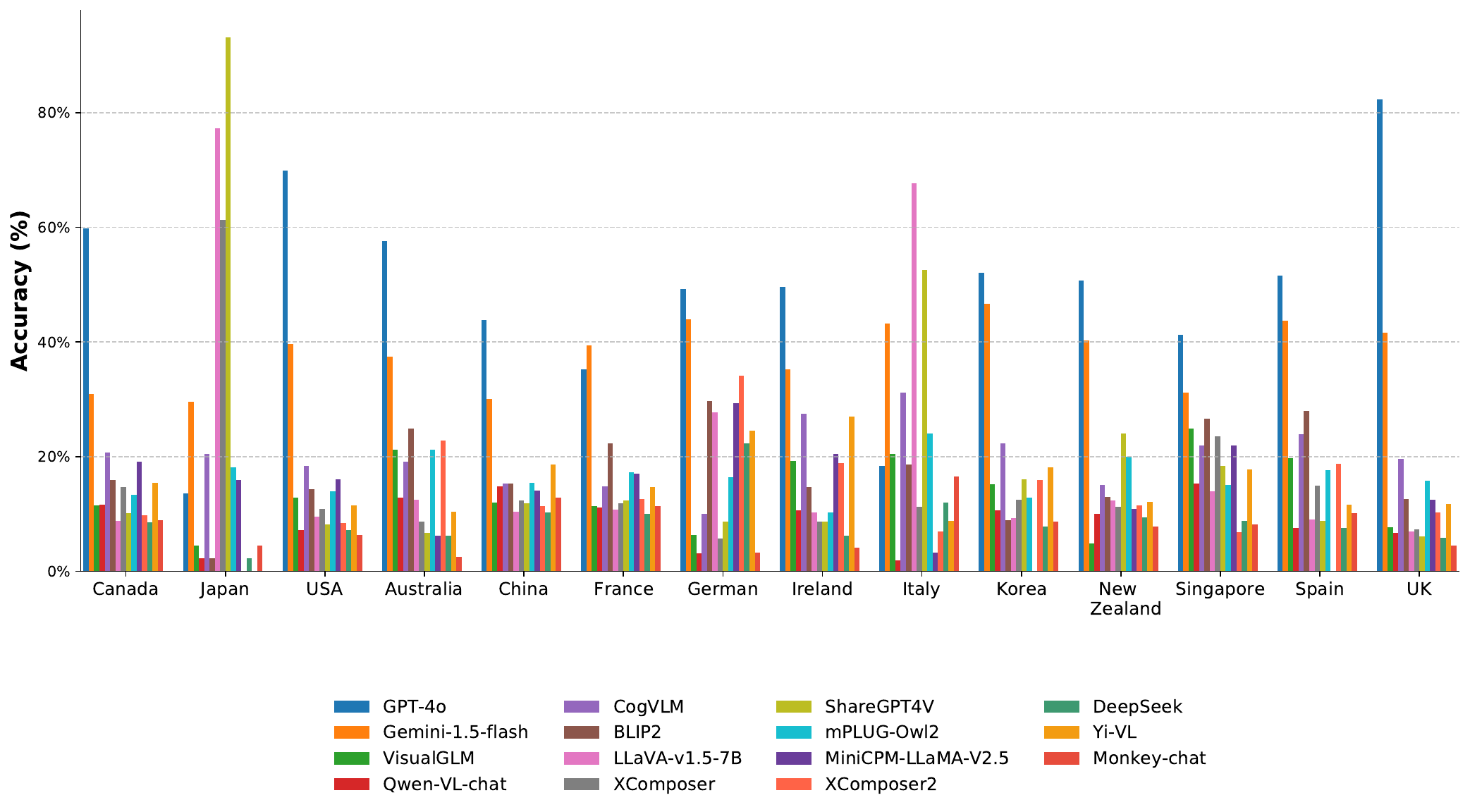}
  \caption{\textbf{Model Accuracy for Multiple-Choice Questions by Country.}}
\label{fig:more_analysis_mcq_country}
\end{figure*}

\begin{figure*}[h]
  \includegraphics[width=\linewidth]{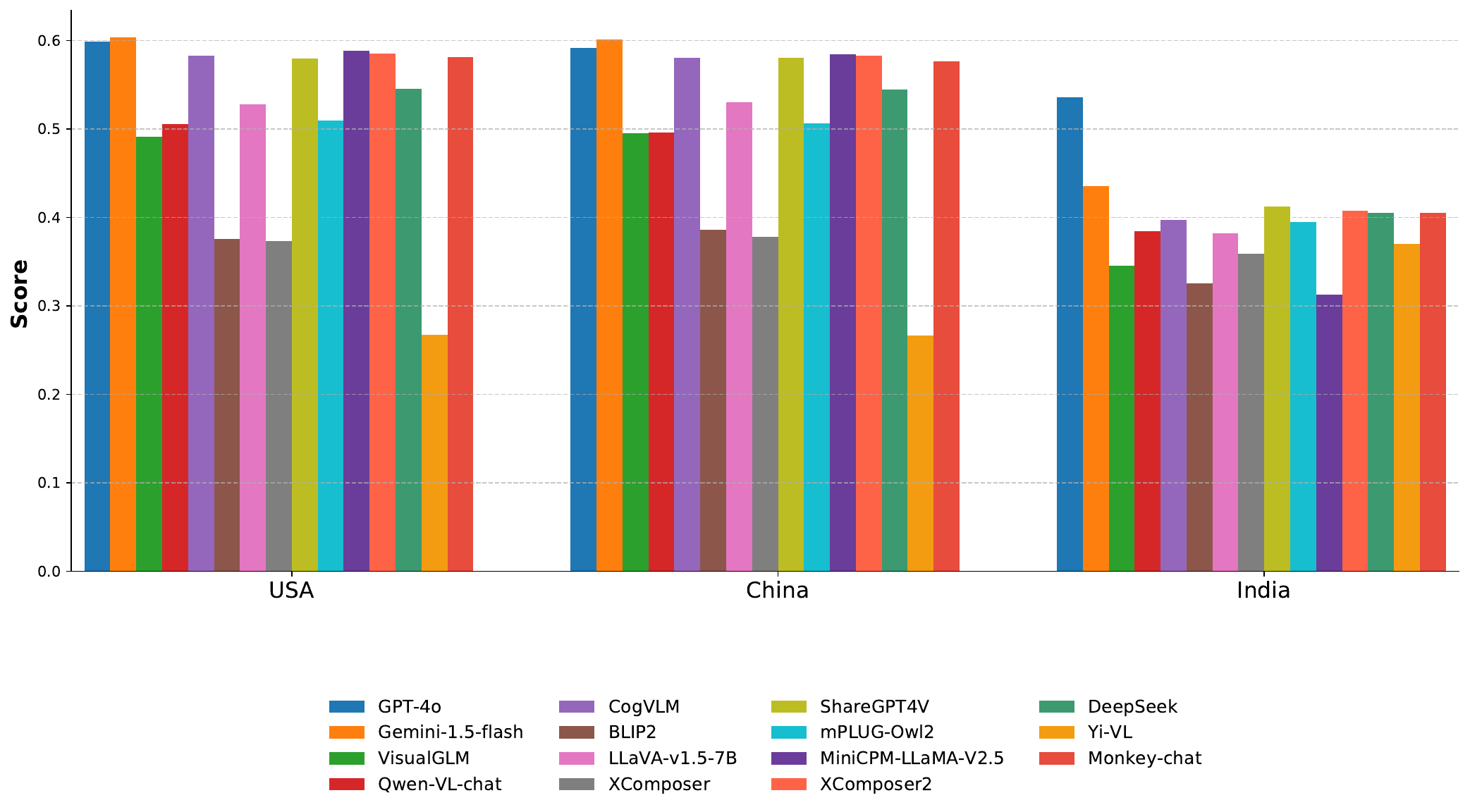}
  \caption{\textbf{Model Score for Question-and-Answer Questions by Country.}}
\label{fig:more_analysis_qa_country}
\end{figure*}

Fig.~\ref{fig:more_analysis_mcq_country} and Fig.~\ref{fig:more_analysis_qa_country} present the performance of various models across different knowledge domains. For MCQ questions, proprietary models show higher accuracy in Driving Techniques and Defensive Driving compared to the other categories. In contrast, most open-source models underperform in these areas, indicating that proprietary models may possess better common-sense knowledge.
For QA questions, the Signs \& Signals category achieves a significantly higher average score than the other categories, although Yi-VL performs poorly in this area. Scores for Driving Techniques and Defensive Driving also surpass those for Laws \& Regulations, highlighting that models tend to perform better in more application-oriented categories than in regulatory ones.

\begin{figure*}[h]
  \includegraphics[width=\linewidth]{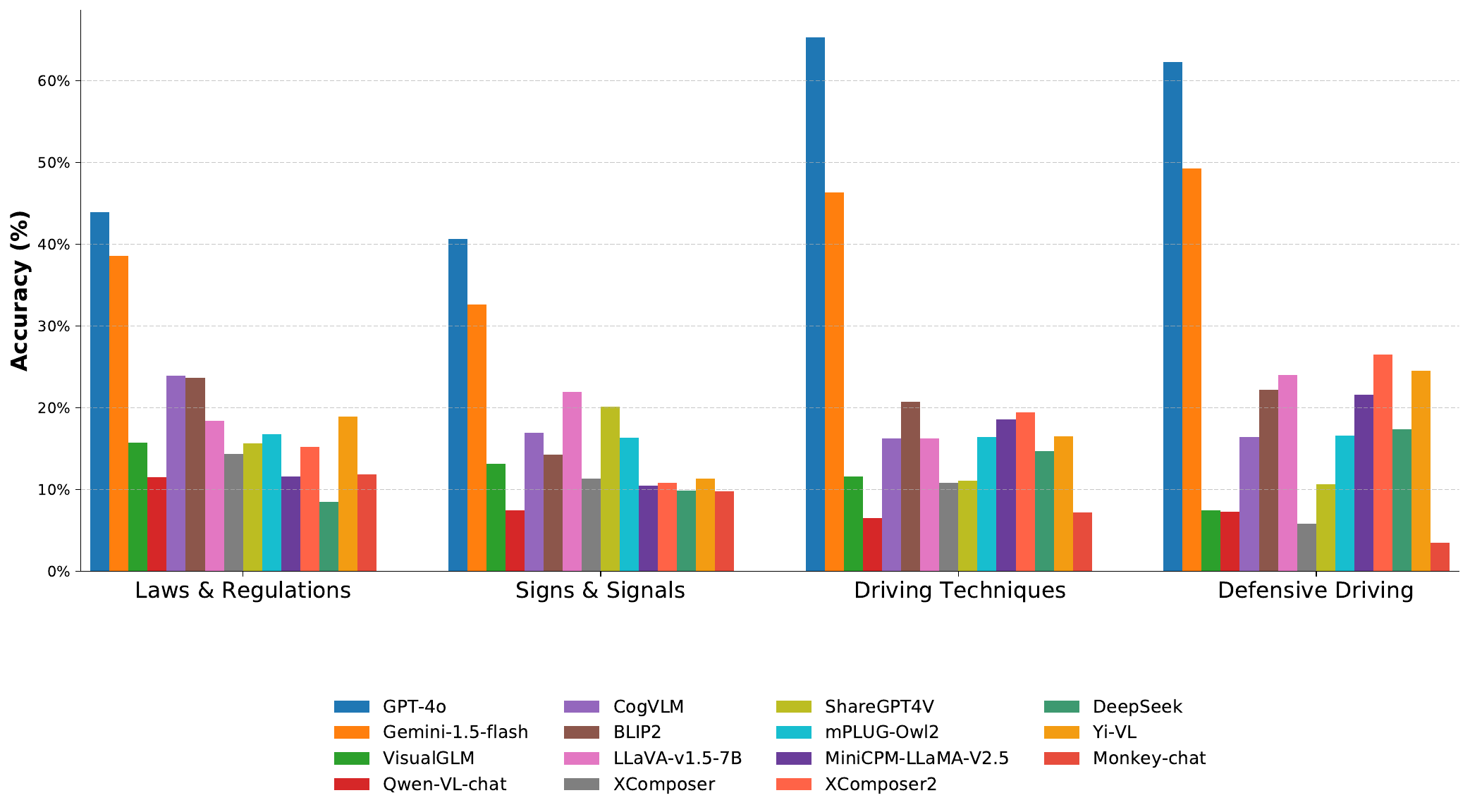}
  \caption{\textbf{Model Accuracy for Multiple-Choice Questions by Knowledge Coverage.}}
\label{fig:more_analysis_mcq_knowledge}
\end{figure*}

\begin{figure*}[h]
  \includegraphics[width=\linewidth]{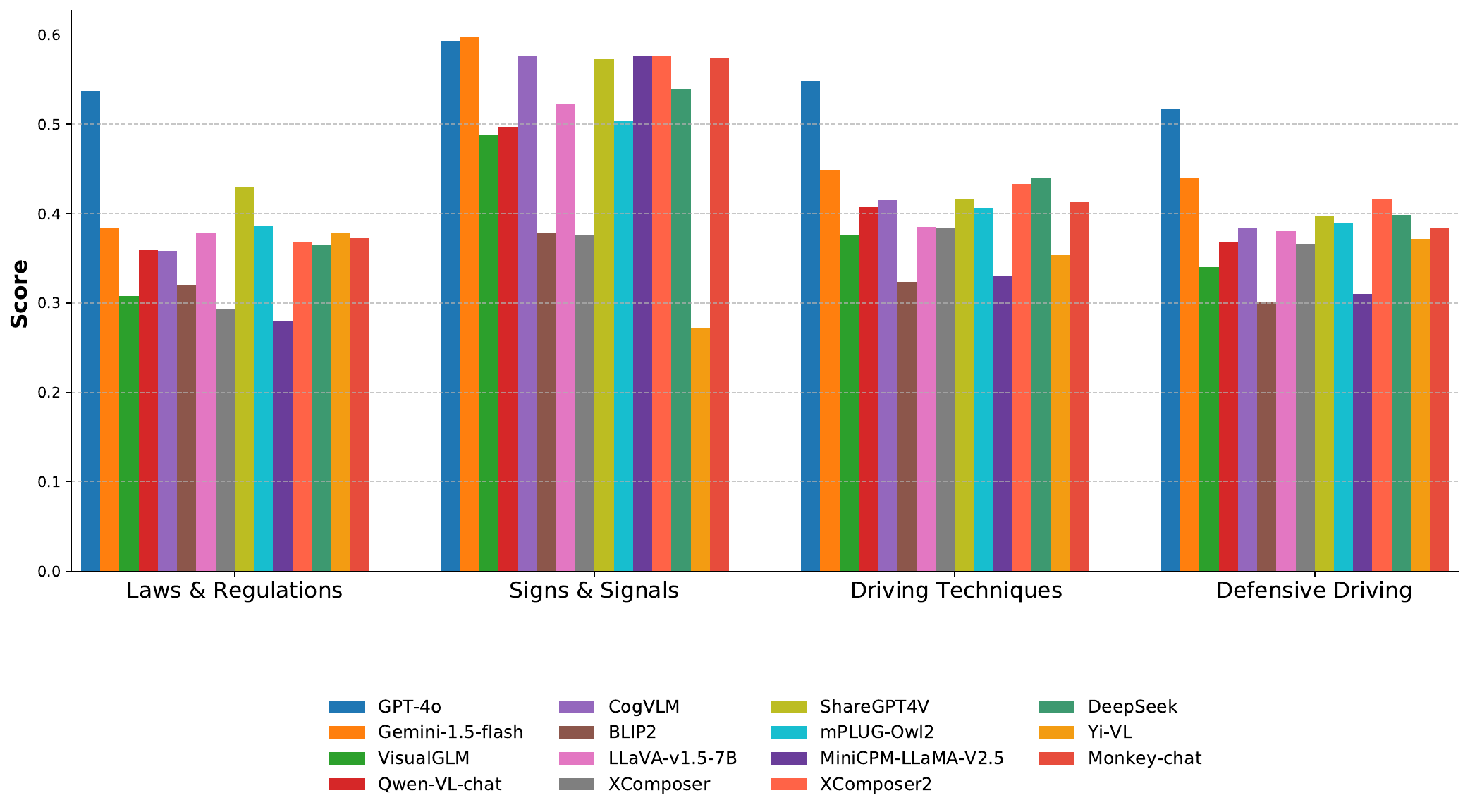}
  \caption{\textbf{Model Score for Question-and-Answer Questions by Knowledge Coverage.}}
\label{fig:more_analysis_qa_knowledge}
\end{figure*}

\end{document}